\pdfoutput=1

\documentclass[11pt]{article}

\usepackage[preprint]{acl}

\usepackage{times}
\usepackage{latexsym}
\usepackage{hyperref} 
\usepackage[T1]{fontenc}

\usepackage[utf8]{inputenc}

\usepackage{microtype}

\usepackage{inconsolata}

\usepackage{graphicx}

\usepackage{amssymb}
\usepackage{amsbsy}
\usepackage{amsmath}
\usepackage{amsthm}
\usepackage{latexsym}
\usepackage{subcaption}
\usepackage{wrapfig}
\usepackage{float}
\usepackage{graphicx}
\usepackage{tabularx}
\usepackage{ulem}
\usepackage[font=small]{caption}
\usepackage{algorithm}
\usepackage{algorithmicx, algpseudocode}
\usepackage{mathtools}
\usepackage{arydshln}
\usepackage{titlesec}
\usepackage{lineno}

\usepackage{xcolor}         
\definecolor{myblue}{RGB}{33, 48, 163}
\definecolor{myred}{RGB}{192, 0, 0}
\usepackage{CJKutf8}
\usepackage{graphicx}

\usepackage{booktabs}

\newcommand{\cc}[1]{\begin{CJK*}{UTF8}{gbsn}#1\end{CJK*}}
\newcommand{\fe}{FinEval}

\linenumbers     
\nolinenumbers   
\title{FinEval: A Chinese Financial Domain Knowledge Evaluation Benchmark for Large Language Models}

\author{Xin Guo$^1$  Haotian Xia$^2$  Zhaowei Liu$^1$ Hanyang Cao$^1$ Zhi Yang$^1$ Zhiqiang Liu$^1$ \\ Sizhe Wang$^1$ Jinyi Niu$^3$ Chuqi Wang$^2$ Yanhui Wang$^4$ Xiaolong Liang$^5$ Xiaoming Huang$^5$ Bing Zhu$^6$ \\Zhongyu Wei$^3$ Yun Chen$^1$ Weining Shen$^2$ Liwen Zhang$^1$}

\author{Xin Guo\textsuperscript{1}$^{\dagger}$,
Haotian Xia\textsuperscript{2}$^{\dagger}$,
Zhaowei Liu\textsuperscript{1}$^{\dagger}$,
Hanyang Cao\textsuperscript{1},
Zhi Yang\textsuperscript{1}, 
Zhiqiang Liu\textsuperscript{1},\\
{\bf Sizhe Wang\textsuperscript{1},
Jinyi Niu\textsuperscript{3},
Chuqi Wang\textsuperscript{2},
Yanhui Wang\textsuperscript{4},
Xiaolong Liang\textsuperscript{5},}\\
{\bf Xiaoming Huang\textsuperscript{5},
Bing Zhu\textsuperscript{6},
Zhongyu Wei\textsuperscript{3},
Yun Chen\textsuperscript{1},
Weining Shen\textsuperscript{2}$^*$,
Liwen Zhang\textsuperscript{1}$^*$}\\
\textsuperscript{1}Shanghai University of Finance and Economics, 
\textsuperscript{2}University of California, Irvine\\
\textsuperscript{3}Fudan University, 
\textsuperscript{4}Ping An Technology (Shenzhen) Co., Ltd.\\
\textsuperscript{5}Tencent YouTu Lab, 
\textsuperscript{6}HSBC Lab, HSBC, Shanghai, China\\
  \texttt{\small\{zhang.liwen\}@shufe.edu.cn},
  \texttt{\small\{weinings\}@uci.edu}
}

\begin{document}
\maketitle

\def\thefootnote{*}\footnotetext{Corresponding authors.}\def\thefootnote{\arabic{footnote}}
\def\thefootnote{$\dagger$}\footnotetext{These authors contributed equally to this work.}\def\thefootnote{\arabic{footnote}}

\begin{abstract}
Large language models have demonstrated outstanding performance in various natural language processing tasks, but their security capabilities in the financial domain have not been explored, and their performance on complex tasks like financial agent remains unknown. This paper presents \fe, a benchmark designed to evaluate LLMs' financial domain knowledge and practical abilities. The dataset contains 8,351 questions categorized into four different key areas: Financial Academic Knowledge, Financial Industry Knowledge, Financial Security Knowledge, and Financial Agent. Financial Academic Knowledge comprises 4,661 multiple-choice questions spanning 34 subjects such as finance and economics. Financial Industry Knowledge contains 1,434 questions covering practical scenarios like investment research. Financial Security Knowledge assesses models through 1,640 questions on topics like application security and cryptography. Financial Agent evaluates tool usage and complex reasoning with 616 questions. \fe~has multiple evaluation settings, including zero-shot, five-shot with chain-of-thought, and assesses model performance using objective and subjective criteria. Our results show that Claude 3.5-Sonnet achieves the highest weighted average score of 72.9 across all financial domain categories under zero-shot setting. Our work provides a comprehensive benchmark closely aligned with Chinese financial domain. The data and the code are available at \href{https://github.com/SUFE-AIFLM-Lab/FinEval}{https://github.com/SUFE-AIFLM-Lab/FinEval}
\end{abstract}

\section{Introduction}
With the development of the financial industry, its integration with large language models has become increasingly close. The financial sector needs large language models to process massive amounts of data, predict market trends, and optimize decision-making processes, thereby helping financial institutions enhance efficiency and reduce risks. This integration requires large models to possess critical capabilities in areas such as financial academic knowledge, industry knowledge, financial security, and financial agents. Financial academic knowledge necessitates that models have a foundational understanding of finance, serving as the baseline for applying large language models in the financial domain. Financial industry knowledge considers language interactions in practical application scenarios, requiring large models to have generalization capabilities across different contexts. Financial security involves various aspects of privacy for individuals and enterprises, which is a priority for the financial industry. Meanwhile, financial agents represent complex tasks within financial scenarios, involving numerous terms and methods that make them difficult for ordinary people to navigate.

There are several well-established benchmarks for evaluating English and Chinese foundation models, such as MMLU~\citep{hendrycks2021measuring}, BIG-bench~\citep{srivastava2022beyond} and GAOKAO-Bench ~\citep{zhang2023evaluating}. Nevertheless, there are some significant drawbacks to these benchmarks for financial tasks: they only cover a small portion of the financial scenarios, and are not widely applicable in real-world financial circumstances. In addition, there are other benchmarks specifically designed to focus on advanced LLMs abilities that become apparent as the scale increases, such as hard math problem-solving~\citep{2021_be83ab3e}, and coding~\citep{chen2021evaluating}. Additionally, there are financial-specific benchmarks such as FinQA~\citep{chen2021finqa}, FinanceIQ~\citep{financeiq} and CFLUE~\cite{zhu2024benchmarking}. 
Although these benchmarks contribute differently to financial tasks, they are all hindered by their limited applicability in real-world situations, narrow scopes, and inability to adequately capture the complexities of financial reasoning. As mentioned in~\citet{he2024emergedsecurityprivacyllm}, with the rapid development of LLMs, they are gradually acquiring the ability to handle complex tasks, but there are still privacy and security issues.~\citet{ding2024largelanguagemodelagent} indicates that LLMs still face challenges in managing complex trading tasks in the financial domain. Therefore, for tasks like financial security and financial agent, which are more closely integrated with real-world financial scenarios and require higher standards, appropriate datasets are needed to assess LLM capabilities. Financial security emphasizes that large models must ensure the protection of personal information and cybersecurity in real-world financial applications. Meanwhile, financial agent highlights the need for large models to possess strong information processing and reasoning capabilities in the complex and dynamic financial market, as well as the ability to flexibly use various financial tools to decompose and solve complex financial tasks. Therefore, evaluating LLMs' capabilities in financial security and financial agent tasks is crucial for the financial domain.

\begin{figure*}[!h]   
    \centering
    \includegraphics[width=\textwidth]{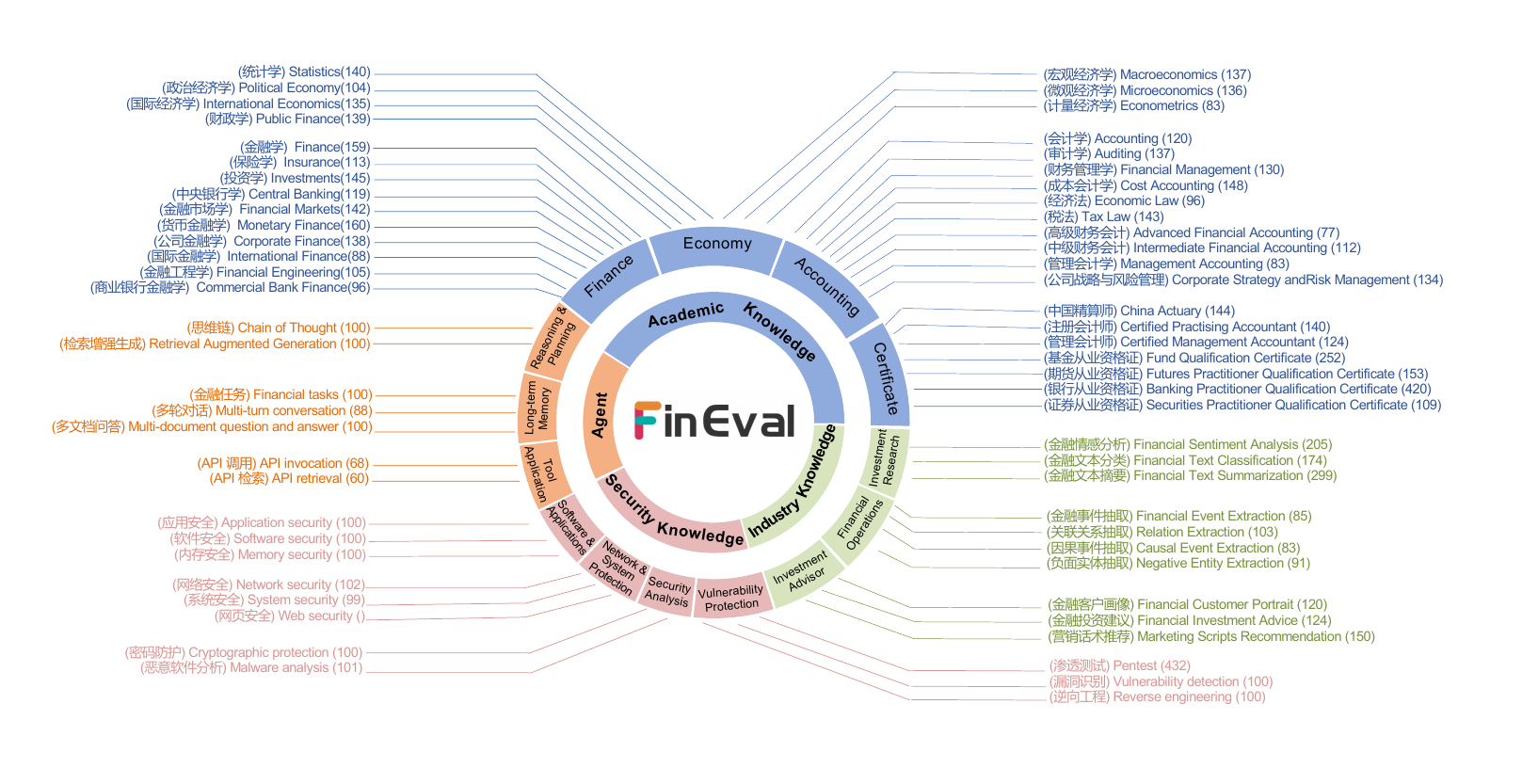}
    \caption{FinEval is divided into four parts:Financial Academic Knowledge, Finance Industry Knowledge, Financial Security Knowledge and Financial Agent. The number of each sub-dataset is indicated after the corresponding name.}.
    \label{fig:scenario}
    \vspace{-2pt}
\end{figure*}

We introduce \fe, an extensive benchmark designed to evaluate the practical capabilities of large language models in the financial domain, with a particular focus on financial security and financial agent tasks within the Chinese context. The dataset contains 8,351 questions distributed divided into major domains: Financial Academic Knowledge, Financial Industry Knowledge, Financial Security Knowledge and Financial Agent, as illustrated in Figure \ref{fig:scenario}. Financial Academic Knowledge comprises 4,661 multiple-choice questions spanning 34 subjects like finance and accounting, testing the theoretical foundation of models. Financial Industry Knowledge, with 1,434 questions, targets real-world financial practices, covering areas such as investment research. Financial Security Knowledge is assessed through 1,640 questions, covering eleven financial security tasks, including Security Analysis, Vulnerability Protection, etc. These questions evaluate the comprehensive capabilities of large language models in terms of security from multiple dimensions. Finally, Financial Agent consists of 616 questions, assessing the performance of large language models under complex information in real financial markets across three major dimensions and seven tasks. Our experiments will evaluate various models, such as Claude 3.5-Sonnet, GPT-4o, Qwen2.5-72B-Instruct, and XuanYuan3-70B-Chat, to demonstrate their capabilities in various financial tasks. These models were assessed
through zero-and few-shot standard prompting, as well as chain-of-thought prompting(\citep{wei2022chain}).

In a series of experiments conducted under a zero-shot setting to evaluate LLM performance on financial knowledge, Claude 3.5-Sonnet performed the best among over 19 models, with the highest weighted average score (72.9). GPT-4o also demonstrated strong capabilities, particularly in financial security, achieving a notable score of 81.8. In addition, the open-source model Qwen2.5-72B-Instruct outperformed Claude 3.5-Sonnet in financial security and matched GPT-4o with a score of 81.8, making it a highly competitive model in this domain. However, this level of accuracy and similarity implies that there is still significant room for improvement in the field of finance for all LLMs. 
\par In summary, our main contributions include:
\begin{itemize}
\item[$\bullet$] We introduced Financial Security Knowledge and Financial Agent based on academic and industry knowledge in finance, creating the first comprehensive dataset for evaluating financial security and agent tasks in the financial domain. The emergence of FinEval addresses the shortcomings of existing financial evaluation benchmarks, providing a comprehensive and in-depth assessment system for evaluating large language models in the financial sector.
\item[$\bullet$] Our work innovatively adds a comparison of the capabilities of large models with those of ordinary individuals and experts in the financial domain, providing a valuable reference for the study of large models' capabilities in finance. The average result shows that large models have surpassed the level of ordinary individuals (30.1) in financial capabilities, but there is still a gap compared to financial experts (85.9), indicating that there is room for improvement in the capabilities of large models within specialized fields.
\item[$\bullet$] Our dataset includes financial academic knowledge derived from publicly accessible mock exam questions, as well as financial industry knowledge compiled and totally rewritten by professionals in the financial field from various publicly available financial websites. Financial security knowledge is adapted from SecEval~\citep{li2023seceval} and developed in collaboration with domain experts with over five years of work experience. The questions for financial agents are manually created by finance experts. Answers are provided by GPT-4o and have undergone multiple rounds of review by financial experts. To better benefit the research community, our dataset will be made publicly available.
\end{itemize}

\section{Related Work}
\noindent\textbf{General Benchmark}\hspace{1em} Current general benchmarks primarily focus on conventional tasks such as natural language understanding, text generation, logical reasoning, programming skills, professional knowledge, and multi-turn dialogues. These benchmarks rarely address tasks in the financial domain, particularly the security issues that are highly valued in finance and complex tasks like those involving agents. There are several well-established benchmarks for evaluating English and Chinese foundation models, including MMLU ~\citep{hendrycks2021measuring}, HELM~\citep{liang2022holistic}, AGIEval ~\citep{zhong2023agieval}, CLUE ~\citep{xu-etal-2020-clue}, and C-Eval ~\citep{huang2023c}. Other benchmarks focus on large language models' advanced abilities, like hard math problem-solving ~\citep{2021_be83ab3e} and coding ~\citep{chen2021evaluating}, which become more apparent as model scale grows. TruthfulQA ~\citep{lin2021truthfulqa} measures the authenticity of language models when answering questions. BIG-bench ~\citep{srivastava2022beyond} evaluates language models across various domains. CBLUE~\citep{zhang2021cblue} is a collection of language understanding tasks in the biomedical field, including named entity recognition and information extraction. GAOKAO-Bench~\citep{zhang2023evaluating} gathers questions from the Chinese Gaokao examination to evaluate the language comprehension and logical reasoning abilities of LLMs. Similarly, AGIEval~\citep{zhong2023agieval}  assess the performance of foundation models on human-centric standardized exams, such as college entrance exams. 

\noindent\textbf{Financial Benchmark}\hspace{1em}However, the availability of benchmarks specifically catering to the financial domain remains limited, and current research mainly focuses on the financial academic and financial industry sectors. FLUE~\citep{shah2022flue}, ConvFinQA~\citep{chen2022convfinqa}, BBT-CFLEB~\citep{lu2023bbt}, and FinQA ~\citep{chen2021finqa} in the English domain all focus solely on knowledge-based question answering. In the Chinese domain, FinanceIQ~\citep{chen2021finqa} also emphasizes knowledge-based questions. CFLUE ~\citep{zhu2024benchmarking} provides questions and NLP tasks related to Chinese financial knowledge, but its actual evaluation tasks are still limited to the knowledge level and do not include important topics such as financial security, nor do they delve into more complex agent-related tasks designed for financial business scenarios. Additionally, there is no comparison between large models and ordinary individuals or experts in the financial domain, making it difficult to accurately assess the true capabilities of large models in financial scenarios.
\section{\fe~Benchmark}
\subsection{Overiew}
\label{sec:design}

\begin{figure*}[ht]
    \centering
\includegraphics[width=1\textwidth]{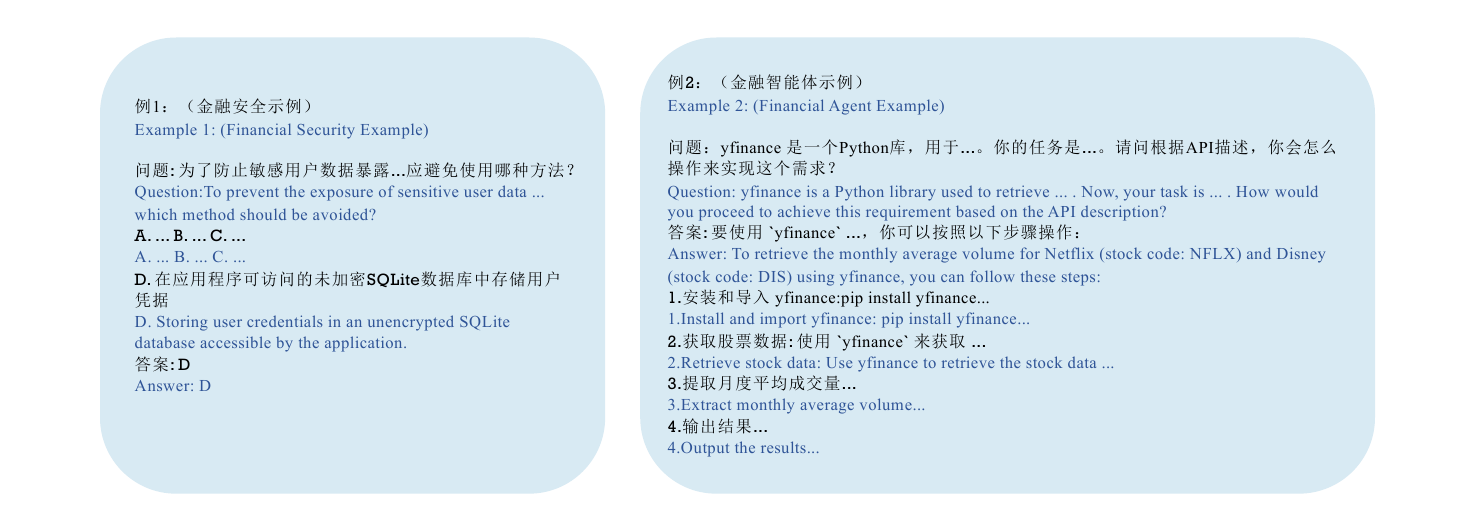}
    \caption{Examples of financial security and financial agent. For better readability, the English translation is displayed below the corresponding Chinese text. Additional examples can be found in Appendix \ref{sec:example}. }
    \label{fig:subjectivequestion}
\end{figure*}

We introduce \fe, a benchmark specifically designed for evaluating large models in the Chinese financial domain. Building on academic and industry knowledge, we further upgrade our focus to address important security and agent tasks in real-world applications within the financial sector.

Financial academic knowledge and financial industry knowledge encompass the fundamental concepts of the financial domain, particularly the subject-specific questions in academic knowledge and various investment recommendations or other specific tasks in industry knowledge. Financial security knowledge is crucial for large models in the financial sector, as it involves all aspects related to user or enterprise information. Large models must possess robust security capabilities to address the various challenges faced by the financial industry. Financial agent tasks involve complex decisions and operations that go beyond simple information processing, requiring large models to have a deep understanding of financial data and the ability to analyze decisions.

As for the question type, financial academic knowledge and security knowledge datasets primarily consist of multiple-choice questions. The multiple-choice questions follow a format similar to that in ~\citet{hendrycks2021measuring}. In financial industry knowledge, we differentiate between semi-open questions (where the answers typically consist of a few words or phrases, or are selected from specific options) and open questions (which require long text responses) through objective short-answer questions and subjective open-ended questions. For the financial agent dataset, the questions are designed as open-ended questions focused on agent-specific tasks. The answers to these open-ended questions are usually long text passages that encompass various outputs, such as task steps, strategies, and results, making them more suitable for complex financial scenarios. Examples of all questions can be found in Appendix \ref{sec:example}.

\subsection{Data Collection}
\label{sec:collect}
\subsubsection{Data source and data quality}
\par Our financial academic knowledge data mainly comes from publicly accessible mock exams and adaptations of questions from certification exams or printed textbooks. The financial industry knowledge data is collected and adapted from various financial websites. All of this data is gathered and adapted by professionals in the financial field, ensuring there are no copyright or other issues. The financial security knowledge data is adapted from SecEval, with the adaptation and annotation work completed by financial security experts with over five years of work experience. Similarly, the financial agent data is produced by these experts in finance, using GPT-4o for answer generation and undergoing multiple rounds of review by them.
\par In terms of data quality, the dataset is collected and adapted by eight postgraduate students with backgrounds in statistics and finance. Three financial experts, specializing in evaluation logic and content, strictly adhere to data quality requirements and are responsible for data quality checks. They manually select and filter questions based on multiple dimensions, including content, direction, logic, and difficulty. After the data adaptation and annotation are completed, the three quality checkers review all the data. Only when all three quality checkers reach a consensus on all aspects of the data is it retained; otherwise, it needs to be re-adapted or annotated. Similarly, the financial security data is adapted and reviewed by three financial security experts from the collaborating organization. As a result, a high-quality FinEval dataset is obtained.

\subsubsection{Data Processing}
The academic knowledge multiple-choice questions in FinEval primarily consist of PDF files, with most sourced from exercise sets in various textbooks, mock exams from different certifications, and past exam questions. All questions in the financial academic knowledge have been processed and refined to include only four options. The multiple-choice questions for financial security knowledge are handled in a similar manner, with financial experts also retaining only four options. The objective short-answer and subjective open-ended questions in financial industry knowledge, as well as the complex open-ended questions in financial agent, are all answered by GPT-4o and reviewed by domain experts. All of the above questions are ultimately converted into a structured format.

 For subjects involving mathematical formulas in financial academic knowledge, we convert them into standard \LaTeX~format, which is a typesetting system commonly used for creating high-quality documents, particularly in academic and technical fields. \LaTeX~allows us to express mathematical expressions directly using text format.
Approximately 100 questions were handled for each subject. Examples can be found in Appendix \ref{sec:example}

\subsection{Statistics}
\par The questions in \fe~contains four parts: Financial Academic Knowledge consists 4661 questions and 34 distinct subjects, which are subsequently classified into broader categories, including Finance, Economy, Accounting, and Certificate. Financial Industry Knowledge consists of 1434 questions and 10 specific directions, which are further categorized into three specific scenarios: Investment research, Investment Advisor, and Financial Operations. 
Financial Security Knowledge consists of 11 specific directions with 1640 questions, which consists of four specific scenarios: Software and Application, Network and Systerm Protection, Security Analysis and Vulnerability Protection. 
Finance Agent consists of 7 categories of tasks and 616 detailed tasks in total.The task is divided into 3 different aspects: Reasoning and Planning, Long-term Memory and Tool Application. All the detail tasks and their broader categories can be found in the Appendix \ref{sec:statistic}, as well as the number of questions included in each task.


\section{Experiments}

Our following experiments show the evaluation results of diverse LLMs on \fe~to analyze their performance and provide baselines for future usage of \fe.

\subsection{Experimental Setup}
\label{metrics}

In this section, we will outline the experimental setup utilized to evaluate the performance of LLMs on financial academic knowledge, financial industry knowledge, financial security knowledge and financial agent. To gauge the adaptability of these large language models, we conducted zero-shot and five-shot with Chain of Thought(CoT). Additionally, we provide specific examples on how to present the prompts in Appendix \ref{sec:example}. Due to limitations related to funding and other factors, we extracted 20\% of the total data as the test set for the evaluation of the final results.

We selected accuracy as the metric for multiple-choice questions in financial academic knowledge and financial security knowledge. In financial industry knowledge, we use Rouge-L \citep{lin-2004-rouge} as the evaluation metric. Agent tasks are typically more complex, comprehensive, and open-ended, with the outputs of large models being longer and more flexible. In such cases, using Rouge-L or other objective evaluation metrics may not accurately assess the quality of the model's output. Therefore, we introduced GPT-4o as a judge model. The judge model scores the responses based on predefined prompts by analyzing several aspects, including the semantic relevance, coherence, logic, and overall quality of the output in relation to the task requirements. GPT-4o is capable of understanding and evaluating the nuances of longer, more diverse responses, making it well-suited for assessing the quality of outputs in complex, multi-dimensional scenarios. The judge scoring criteria prompt can be found in the Appendix \ref{sec:example}.

As a result, our model evaluation encompasses four types of scenarios: zero-shot prompting, five-shot prompting, zero-shot CoT prompting and five-shot CoT prompting. Due to the higher complexity of financial agent tasks, we only conduct evaluations on the financial agent data under the zero-shot prompting setting. 

\subsection{Models}
To achieve a comprehensive understanding of the state of LLMs in the context of the Chinese language, we conducted an evaluation of 19 high-performing LLMs that can process Chinese input. The detailed information about the large models participating in the evaluation can be found in Appendix \ref{sec: models overview}.

\paragraph{Closed-Source Models:}
In the realm of closed-source models, we evaluated six leading, high-performance LLMs provided by three organizations, including GPT-4o~\citep{openai2024gpt4o} and GPT-4o-mini~\citep{openai2024gpt4omini} from OpenAI, Claude 3.5-Sonnet~\citep{Claude3.5} from Anthropic, and Gemini1.5-Flash and Gemini1.5-Pro~\citep{geminiteam2024gemini15unlockingmultimodal} from Google. 

\paragraph{Open-Source Models:}
For open-source models, we evaluated seven mainstream LLMs capable of understanding and generating Chinese, including Baichuan2-13B-Chat~\citep{baichuan2-13b-chat}, 
Yi1.5-9B-Chat~\citep{Yi1.5-9b/34b}, Yi1.5-34B-Chat~\citep{Yi1.5-9b/34b}, 
ChatGLM3-6B~\citep{chatglm3-6b}, GLM-4-9B-Chat~\citep{glm2024chatglm}, 
InternLM2-20B-Chat~\citep{2023internlm}, InternLM2.5-20B-Chat~\citep{2024internlm},
Qwen2.5-7B-Instruct~\citep{qwen2.5-7b}, Qwen2.5-72B-Instruct~\citep{qwen2.5-72b}, etc. 

\paragraph{Financial Domain Models:}
In financial domain, we evaluated five representative LLMs tailored for the financial tasks, including  DISC-FinLLM~\citep{disc-finllm}, FinGPTv3.1~\citep{fingpt}, CFGPT2-7B~\citep{cfgpt2-7b}, XuanYuan2-70B-Chat~\citep{xuanyuan2} and XuanYuan3-70B-Chat~\citep{xuanyuan3}.

\begin{table*}[ht]
\caption{Average zero-shot scores across four evaluated categories. We report the results under zero-shot setting for four categories and one final weighted average:  Financial Academic Knowledge, Financial Industry Knowledge, Financial Security Knowledge and Financial Agent. For the scoring criteria for each section, please refer to Section \ref{metrics}. As for the details of the models involved in the evaluation, you can refer to Table \ref{tab:model-overview} in Appendix \ref{sec: models overview}.}
\label{Average zero-shot scores}
\vspace{10pt}
\resizebox{0.9\textwidth}{!}{
    \begin{tabular}{llccccc}
    \toprule[2pt]
    \textbf{Model} & \textbf{Size} &\textbf{Financial Academic} & \textbf{Financial Industry} & \textbf{Financial Security} & \textbf{Financial Agent} & \textbf{Weighted Average} \\
    \midrule[1pt]
    Claude 3.5-Sonnet &unknown & \textbf{73.9}  & 60.6   & 78.1  & \textbf{79.3}  & \textbf{72.9}     \\
    GPT-4o & unknown &71.5  & \textbf{61.3}  & \textbf{81.8}  & 73.9  & 71.9  \\
    Qwen2.5-72B-Instruct & 72B&69.7  & 54.4  & \textbf{81.8}  & 68.4  & 69.4  \\
    Gemini1.5-Pro &unknown & 68.3  & 60.5  & 77.8  & 72.8  & 69.2  \\
    GPT-4o-mini &unknown & 62.4  & 61.1  & 79.1  & 72.9  & 66.2  \\
    Gemini1.5-Flash & unknown &62.1  & 61.2  & 77.5  & 70.9  & 65.6  \\
    Qwen2.5-7B-Instruct &7B& 62.7  & 48.3  & 71.7  & 66.7  & 62.3  \\
    Yi1.5-34B-Chat &34B& 59.5  & 49.6  & 76.0  & 66.0  & 61.5  \\
    XuanYuan3-70B-Chat &70B& 55.2  & 52.0  & 74.4  & 63.9  & 59.1  \\
    InternLM2.5-20B-Chat &20B& 54.7  & 53.2  & 74.1  & 63.1  & 58.9  \\
    GLM-4-9B-Chat &9B& 54.7  & 53.1  & 73.1  & 60.2  & 58.4  \\
    InternLM2-20B-Chat  &20B& 54.7  & 50.3  & 73.1  & 60.9  & 58.0  \\
    Yi1.5-9B-Chat &9B& 55.0  & 44.7  & 71.4  & 61.1  & 56.9  \\
    XuanYuan2-70B-Chat &70B& 52.8  & 46.6  & 68.0  & 61.7  & 55.4  \\
    CFGPT2-7B &7B& 53.9  & 50.2  & 65.1  & 50.9  & 55.3  \\
    Baichuan2-13B-Chat  &13B& 41.1  & 50.2  & 61.6  & 55.7  & 47.8  \\
    ChatGLM3-6B &6B& 38.9 & 48.6  & 48.2  & 49.6  & 43.2  \\
    DISC-FinLLM &13B& 39.1  & 46.1  & 25.2  & 41.8  & 37.8  \\
    FinGPTv3.1 &6B& 25.3  & 36.1  & 22.7  & 31.2  & 27.1  \\
    \bottomrule[2pt]
    \end{tabular}
}
\centering
\end{table*}

\subsection{Results}
We evaluated the models in four settings: zero-shot, five-shot, zero-shot CoT, and five-shot CoT. However, due to the extensive variety, high difficulty, and complexity of financial agent tasks, we only evaluated financial academic knowledge, financial industry knowledge, and financial security knowledge in the other three settings, with these results available in Appendix \ref{other results}. In the zero-shot setting, we present the results of 19 models participating in the evaluation across the four independent tasks.

Table \ref{Average zero-shot scores} showcases the abilities of 19 models under the setting of zero-shot. Among them, Claude 3.5-Sonnet and GPT-4o has shown outstanding capabilities with an weighted average score exceeding 70 and performing the best in all four task categories. 
When comparing models across three categories, we find that overall, closed-source models outperform open-source models, which in turn outperform models specialized in the financial domain. Among the open-source general and financial models, Qwen2.5-72B-Instruct ranks the highest with an weighted average score of 69.4. Following closely are Qwen2.5-7B-Instruct, Yi1.5-34B-Chat and XuanYuan3-70B-Chat. This demonstrates the outstanding capabilities of these open-source models in the financial domain. Furthermore, it can be observed that general models rank relatively higher compared to financial models, while except XuanYuan3-70B-Chat, other fine-tuned models rank relatively lower, indicating that general models perform better in financial domain and suggesting their superior task generalization abilities. 
In our five-shot CoT setting, Qwen2.5-72B-Instruct performed exceptionally well, ranking first compared to its third-place ranking in the zero-shot setting. The financial large model XuanYuan3-70B-Chat also climbed from ninth place in the zero-shot setting to first place. This trend is also observed in some later-released open-source large models, such as GLM-4-9B-Chat. This indicates that open-source large language models can enhance their performance through prompt optimization guidance.

\begin{table}[ht]
\centering
\caption{Performance comparison across ordinary individuals, experts and LLMs(selected with top 2 model's results within each category).}
\label{contrast analysis}
\resizebox{0.48\textwidth}{!}{
\begin{tabular}{llcccc}
    \toprule[2pt]
    \textbf{Source} & \textbf{Category}& \textbf{Financial Academic} & \textbf{Financial Security}  &\textbf{Average}\\
    \midrule[1pt]
    Human & Ordinary individual & 35.1 & 26 &30.1 \\
     & Experts & \textbf{84.9} & \textbf{86.8} &\textbf{85.9} \\
    \midrule[1pt]
    Closed-source & Claude 3.5-Sonnet & 73.9 & 78.1&76 \\
    & GPT-4o & 71.5 & 81.8 &76.7\\
    \midrule[1pt]
    General & Qwen2.5-72B-Instruct & 69.7 & 81.8&75.8\\
    & Yi1.5-34B-Chat & 59.5 & 76 &67.8 \\
    \midrule[1pt]
    Financial & XuanYuan3-70B-Chat & 55.2 & 74.4 &64.8 \\
    & CFGPT2-7B & 53.9 & 65.1  &59.5\\
    \bottomrule[2pt]
\end{tabular}}
\end{table}

\subsection{Contrast Analysis}
To better evaluate the capabilities of large language models and to make meaningful contributions to model research, we organized a competition between these models, ordinary people, and financial experts. In our dataset, we selected relatively easier multiple-choice questions as competition topics. Considering that some question-and-answer items may be difficult for ordinary individuals and that a limited number of questions would not be representative, we extracted 20\% from the financial academic knowledge and financial security knowledge used for evaluation, totaling 260 questions for testing.
\par For the ordinary participants, we randomly selected three undergraduate students who had no prior exposure to financial or security-related knowledge. For the expert responses, the financial academic knowledge questions were answered by PhD students in finance, while the financial security knowledge questions were answered by other security-related experts with over five years of work experience. To ensure the validity of the results, all participants involved in the testing have not been exposed to any FinEval questions. We present the comparison results between the large language models, ordinary individuals, and financial experts in Table \ref{contrast analysis}. It can be seen that in both the financial academic knowledge and financial security knowledge test results, the performance of closed-source, general, and financial domain models far exceeds that of ordinary individuals, with closed-source models and some general models achieving notably high performance. However, the overall results of the large models still have a gap compared to the experts, with the best models showing nearly a 10\% difference from expert results. Nevertheless, certain models, such as GPT-4o and Qwen2.5-72B-Instruct, have demonstrated capabilities in safety that are very close to expert levels, with a gap of around 5\%, reflecting the current emphasis on safety in large models.

\begin{figure*}
    \centering
    \includegraphics[width=0.75\textwidth]{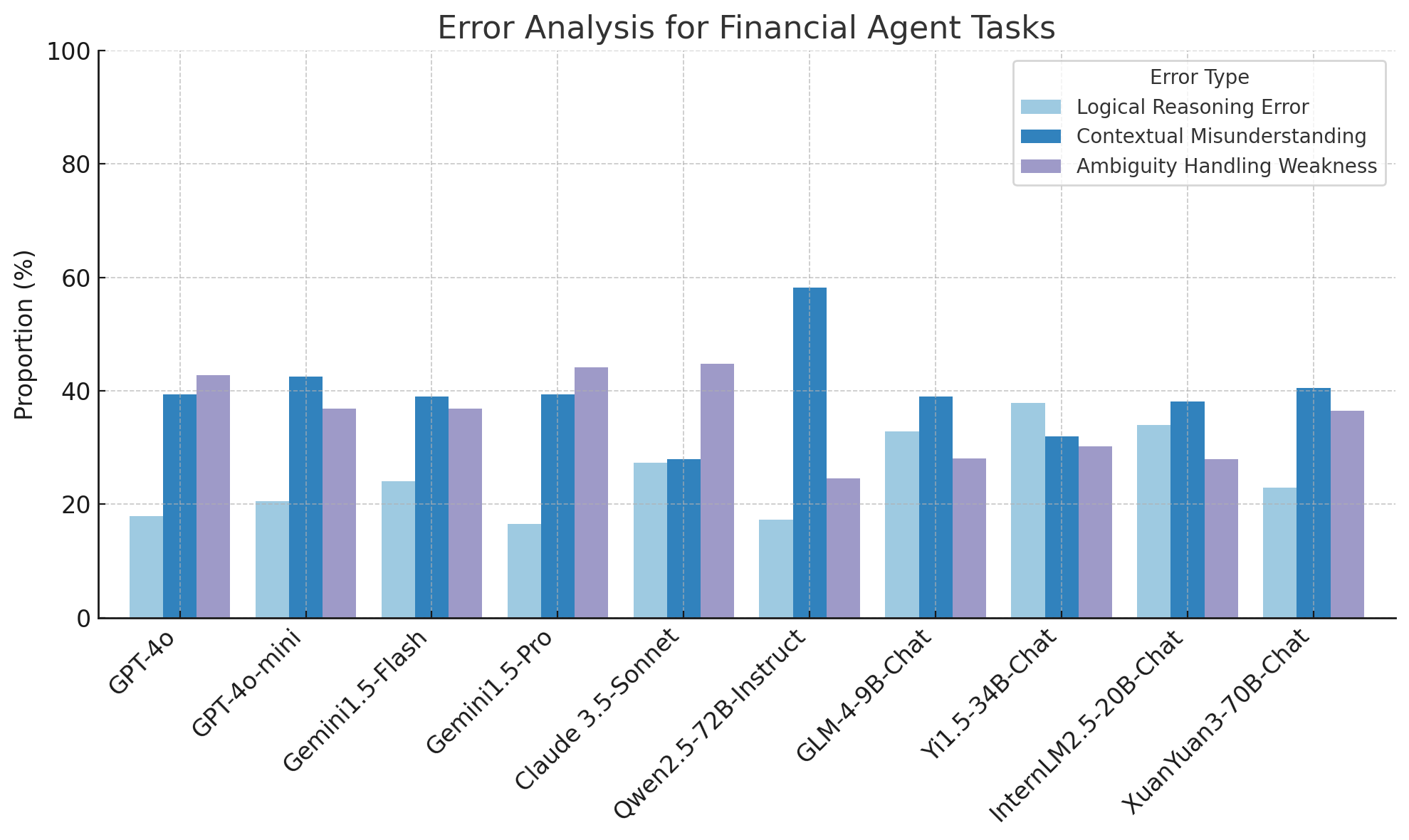}
    \caption{Error analysis results of ten models. Each bar represents the proportion of a specific type of error among all errors made by a particular model. The sum of the values of the three bars for a model equals to 1, representing the total error distribution for that model.} 
    \label{fig:error analysis}
\end{figure*}
\section{Error Analysis}
To further identify the shortcomings of large language models in financial knowledge and tasks, we analyze the errors made by the model during the testing process. Financial Agent includes a series of relatively open-ended, subjective tasks that place high demands on the model, and the model displayed a diverse and rich variety of error types when completing more complex tasks. Three types of errors were identified for answers that did not receive full scores from the evaluation model: Logical Reasoning Error, Contextual Misunderstanding, and Ambiguity Handling Weakness. Logical Reasoning Error occurs when the model fails to draw correct logical conclusions. Contextual Misunderstanding happens when the model misses or misinterprets relevant context. Ambiguity Handling Weakness refers to the model's inability to handle unclear or ambiguous questions properly. Detail examples can be found in Appendix \ref{sec:example}. We selected ten representative models and randomly sampled 27 erroneous responses for each model on each task. The results are shown in Figure \ref{fig:error analysis}. Each bar in the chart represents the proportion of a specific type of error among all errors made by a particular model.

Across all the erroneous responses, the average proportion for open-source models in Logical Reasoning Error, Contextual Misunderstanding, and Ambiguity Handling Weakness are 29.0\%, 41.6\%, and 29.4\%, respectively. In comparison, the corresponding average proportion for closed-source models are 21.3\%, 37.6\%, and 41.1\%. This suggests closed-source models are better at reasoning and understanding long texts, but more prone to semantic issues and hallucinations. This could be due to their larger parameter sizes, which, while improving reasoning and comprehension, also lead to over-interpretation of ambiguous content. Specifically, the results show that open-source models are more likely to encounter Logical Reasoning Errors and Contextual Misunderstanding, while closed-source models are more prone to Ambiguity Handling Weakness. For Logical Reasoning Errors, models struggle with multi-step reasoning tasks, often misusing formulas or making calculation mistakes. In Contextual Misunderstanding, models fail to connect distant parts of the text. For Ambiguity Handling Weakness, models have trouble interpreting vague information, sometimes leading to hallucinations. They often include unnecessary tools or data in their responses, indicating a need for better semantic understanding and relevance filtering.
\section{Conclusion}
This study introduces FinEval, a benchmark for evaluating large language models' capabilities in the financial domain. Unlike previous financial evaluation benchmarks, FinEval delves deeper into financial security and financial agent, covering pressing security issues and complex agent tasks in the financial field, assessing models' security and their ability to handle complex tasks. Our results indicate that Claude 3.5-Sonnet performs the best among the 19 models evaluated, but it still faces challenges with more complex tasks. While it surpasses ordinary individuals, it has not reached the level of human experts. This study illustrates that although large language models have made certain breakthroughs in the financial domain, they still require a more in-depth and detailed understanding to enhance their task generalization capabilities in a wider and more complex diverse financial market environment. As an important benchmark for future research on large language models in the financial field, FinEval provides a structured framework for measuring and improving the capability of large language models, contributing to the development of evaluation benchmarks in the Chinese financial domain.

\section*{Limitations}
Although FinEval has become a relatively comprehensive evaluation benchmark in the financial domain, encompassing a wide range of financial tasks, we acknowledge its limitations. With the ongoing emergence and iteration of data formats such as image, audio, and video, there is an increasing amount of multimodal data in the financial domain, which presents a limitation for the current \fe. This limitation highlights the need to develop a multimodal evaluation dataset as the next focus for \fe. We will recruit more specialized financial personnel to collect additional multimodal data related to the financial domain (especially financial chart data, which is particularly important), to evaluate a broader range of multimodal large language models, ensuring that FinEval  remains a comprehensive benchmark in the financial field.
\section*{Acknowledgments}

This research is supported in part by the National Social Science Fund of China 22BTJ031 (Liwen Zhang). We sincerely appreciate the assistance and guidance provided by Professor Xiao Cao and Associate Professor Min Min from the School of Finance at Shanghai University of Finance and Economics, as well as Shen Wang from the Experimental Center at Shanghai University of Finance and Economics.

\bibliography{FinEval}

\newpage
\appendix

\section{Evaluated Models Overview}
\label{sec: models overview}
We list the models we evaluated in this paper in Table \ref{tab:model-overview}.
\begin{table*}[ht]
\centering
\caption{Models evaluated in this paper. The "Access" column shows whether we have full access to the model weights or we can only access through API. The “Version Date” column shows the release date of the corresponding version of the model we evaluated.}
\label{tab:model-overview}
\vspace{10pt}
\resizebox{0.9 \textwidth}{!}{
\begin{tabular}{llcccc}
    \toprule[2pt]
    \textbf{Category} & \textbf{Model} & \textbf{Creator} & \textbf{Parameter} & \textbf{Access} & \textbf{Version Date}\\
    \midrule[1pt]
    \textbf{Closed-Source}& GPT-4o & OpenAI & undisclosed & API &2024.5\\
    &GPT-4o-mini & OpenAI & undisclosed  & API &2024.7\\
    &Gemini1.5-Flash & Google & undisclosed & API & 2024.5\\
    &Gemini1.5-Pro & Google & undisclosed & API & 2024.5\\   
    &Claude 3.5-Sonnet & Anthropic & undisclosed & API & 2024.3\\
    \midrule[1pt]
    \textbf{Open-Source} &Qwen2.5-7B-Instruct & Alibaba Cloud  & 7B    & Weights &2024.9\\
    &Qwen2.5-72B-Instruct & Alibaba Cloud  & 72B    & Weights &2024.9\\
    &ChatGLM3-6B & Tsinghua \& Zhipu.AI  & 6B    & Weights &2023.10\\
    &GLM-4-9B-Chat & Tsinghua \& Zhipu.AI  & 9B    & Weights &2024.6\\
    &Yi1.5-9B-Chat & 01.AI & 9B   & Weights &2024.5\\
    &Yi1.5-34B-Chat & 01.AI & 34B   & Weights &2024.5\\
    &InternLM2-20B-Chat & Shanghai AI Lab \& SenseTime & 20B  & Weights &2024.1\\
    &InternLM2.5-20B-Chat & Shanghai AI Lab \& SenseTime & 20B  & Weights &2024.8\\
    &Baichuan2-13B-Chat & Baichuan & 13B   & Weights &2023.12\\
    
    \midrule[1pt]
    \textbf{Financial} &XuanYuan3-70B-Chat & Duxiaoman-DI   &   70B    &Weights   &2024.9\\
    &XuanYuan2-70B-Chat & Duxiaoman-DI   &   70B    &Weights   &2024.3\\
    &DISC-FinLLM     & FudanDISC    &    13B    &Weights     &2023.10\\
    &CFGPT2-7B & TongjiFinLab & 7B & Weights &2024.8\\
    &FinGPTv3.1  & AI4Finance-Foundation  &  6B &Weights &2023.10\\

    \bottomrule[2pt]
\end{tabular}}

\end{table*}

\section{Examples for \fe}
\label{sec:example}
\begin{figure*}[!h]
    \centering
\includegraphics[width=1\textwidth]{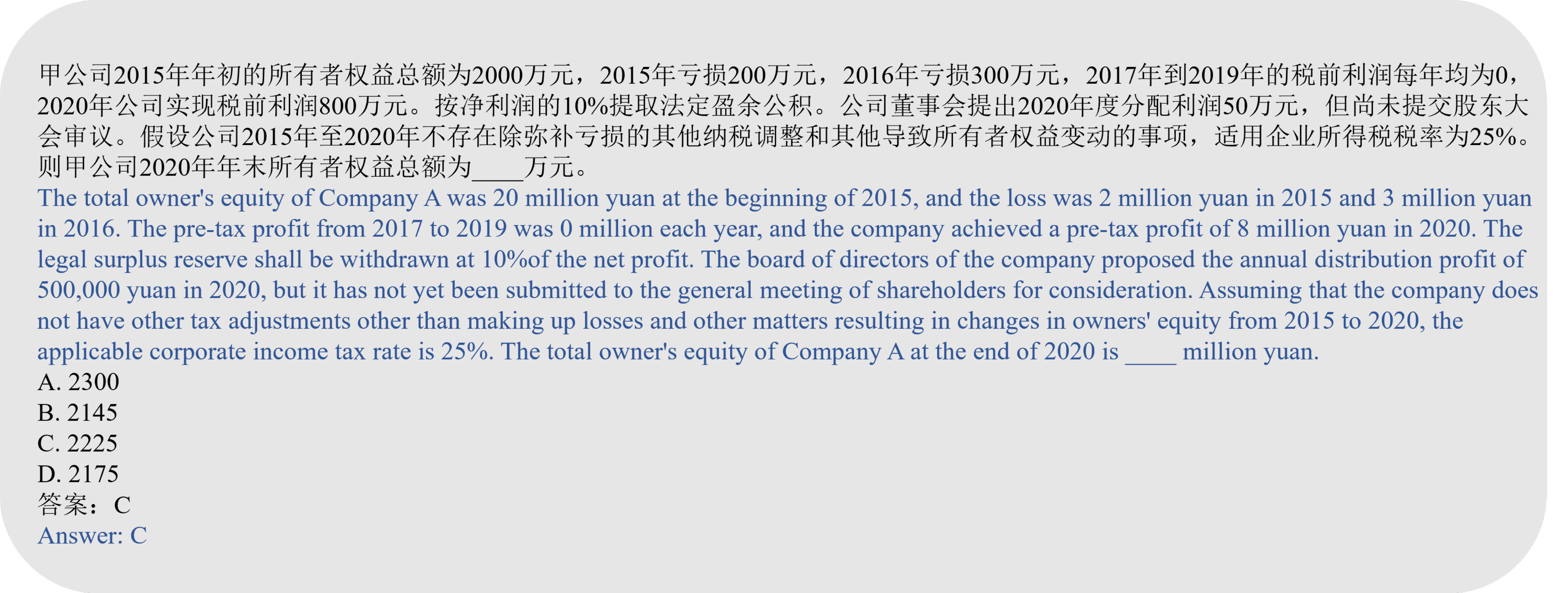}
    \caption{Zero-shot example of multiple-choice questions in Intermediate Financial Accounting. For better readability, the English translation is displayed below the corresponding Chinese text.}
    \label{fig:easyexample}
\end{figure*}

\begin{figure*}[ht]
   \centering \includegraphics[width=1\textwidth]{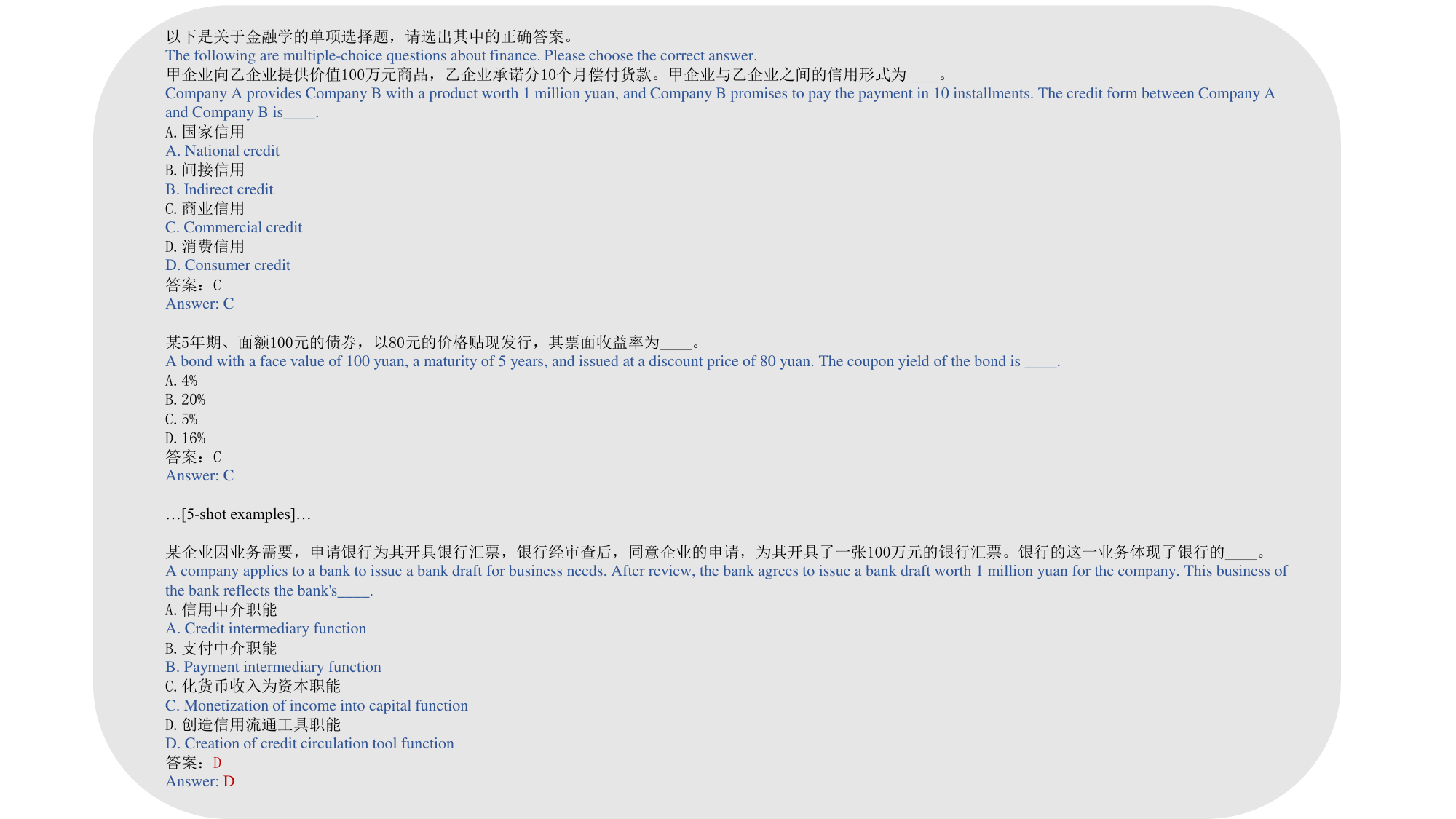}
  \caption{An instance of five-shot evaluation. The red text denotes the response automatically generated by the model, with the preceding text being the input prompt. English translations for the related Chinese text are provided beneath.}
    \label{fig:five_shot}
\end{figure*}

\begin{figure*}[ht]
    \centering \includegraphics[width=1\textwidth]{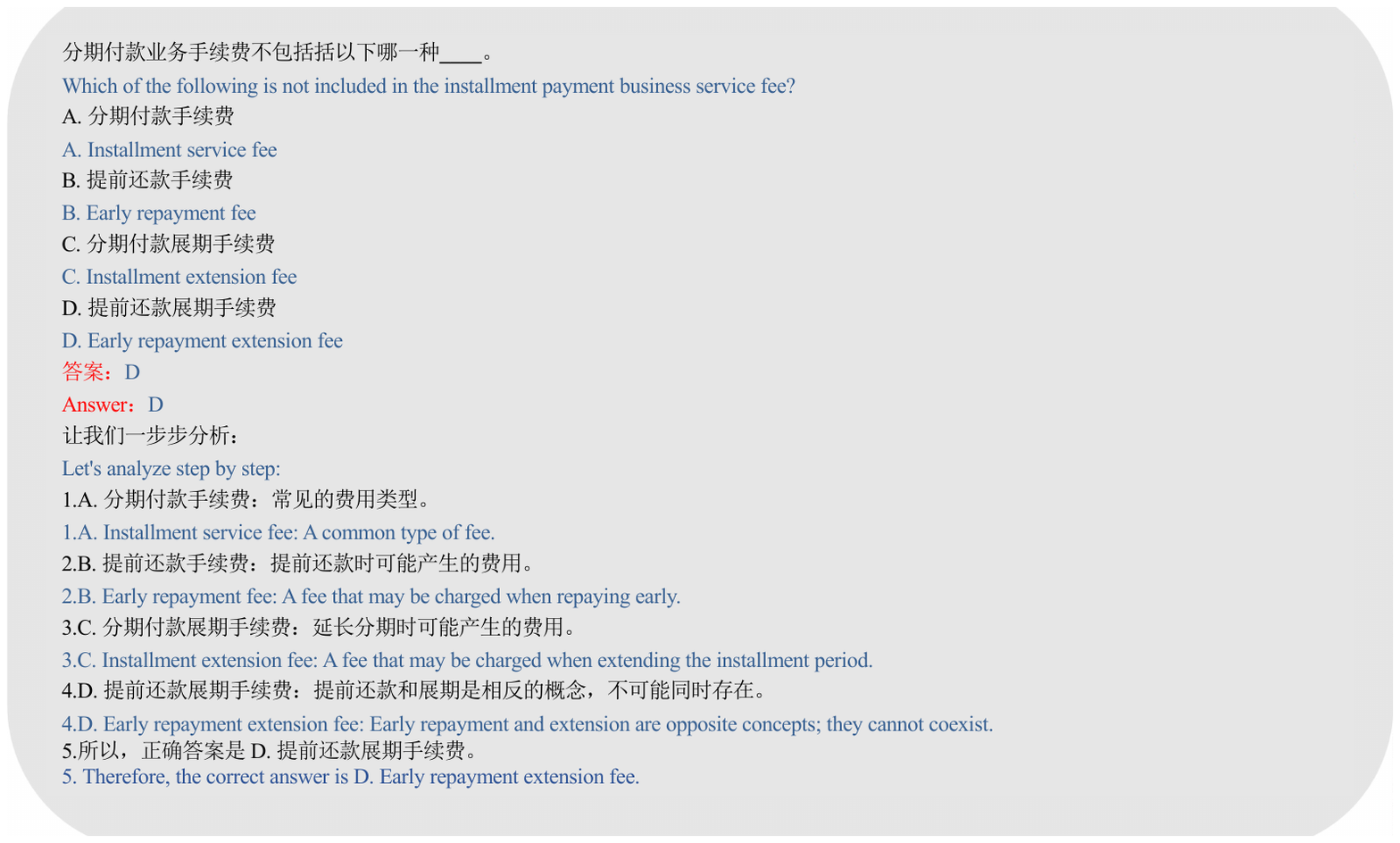}
    \caption{An example of zero-shot CoT evaluation. The red text is followed by the response automatically generated by the model. The English translations for the related Chinese text are provided below.}
    \label{fig:cot_zero_shot}
\end{figure*}

\begin{figure*}[ht]
    \centering \includegraphics[width=1\textwidth]{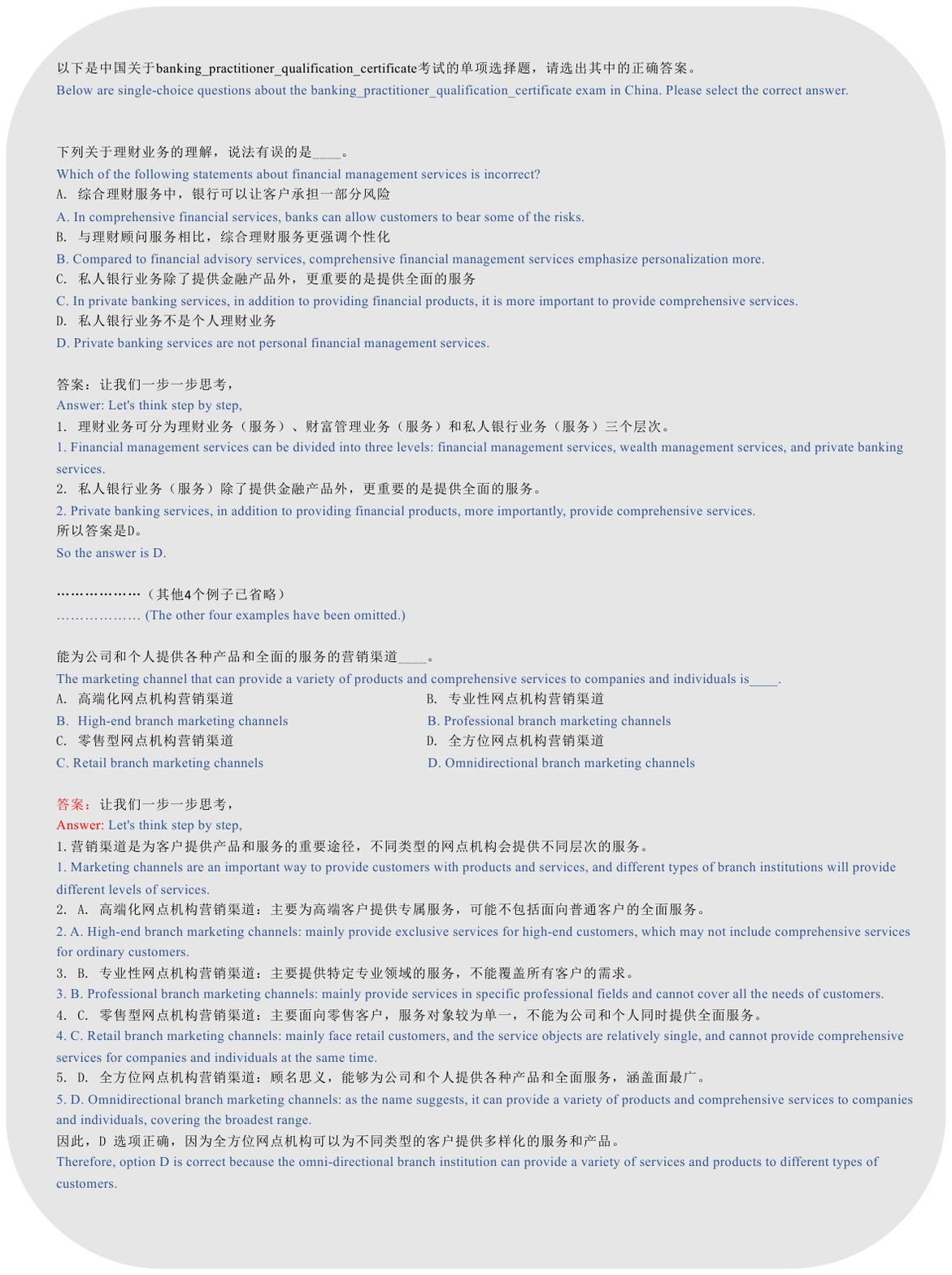}
    \caption{An example of five-shot CoT evaluation. The preceding text serves as an example, for convenience, the other four examples have been omitted. The red text is followed by the response automatically generated by the model. The English translations for the related Chinese text are provided below.}
    \label{fig:cot_five_shot}
\end{figure*}

\begin{figure*}[!h]
    \centering
\includegraphics[width=1\textwidth]{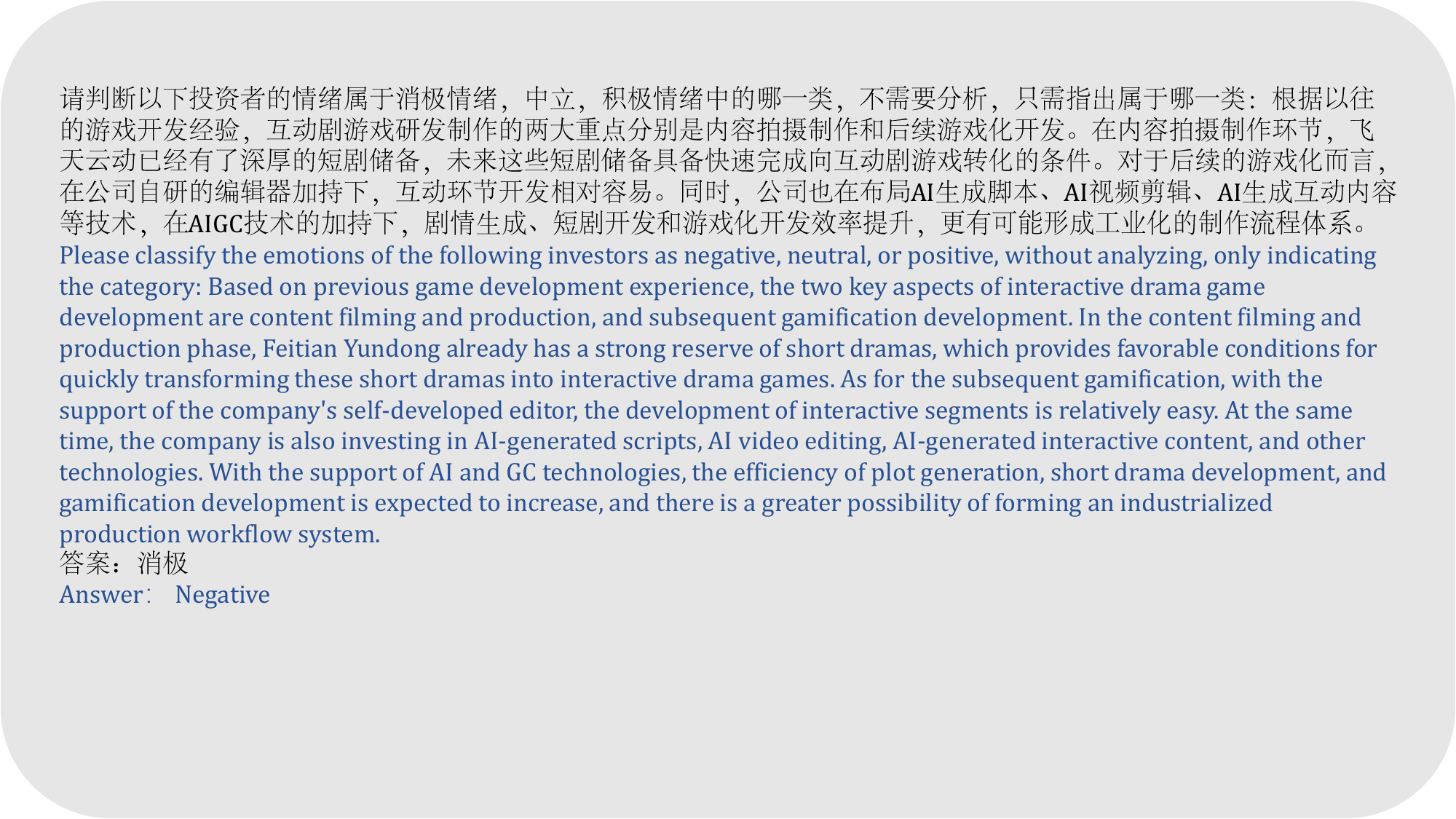}
    \caption{An example of objective short-answer questions in Financial Sentiment Analysis. For better readability, the English translation is displayed below the corresponding Chinese text.}
    \label{fig:objectivequestion}
\end{figure*}

\begin{figure*}[ht]
    \centering
\includegraphics[width=1\textwidth]{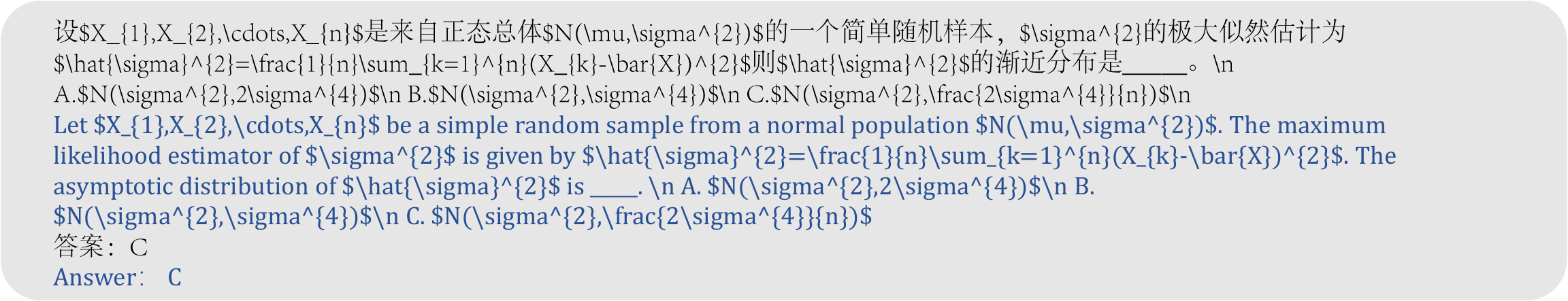}
    \caption{An example of \LaTeX~format in the subject of Statistics, under the category of economy.}
    \label{fig:latex}
\end{figure*}

\begin{figure*}[ht]
    \centering \includegraphics[width=1\textwidth]{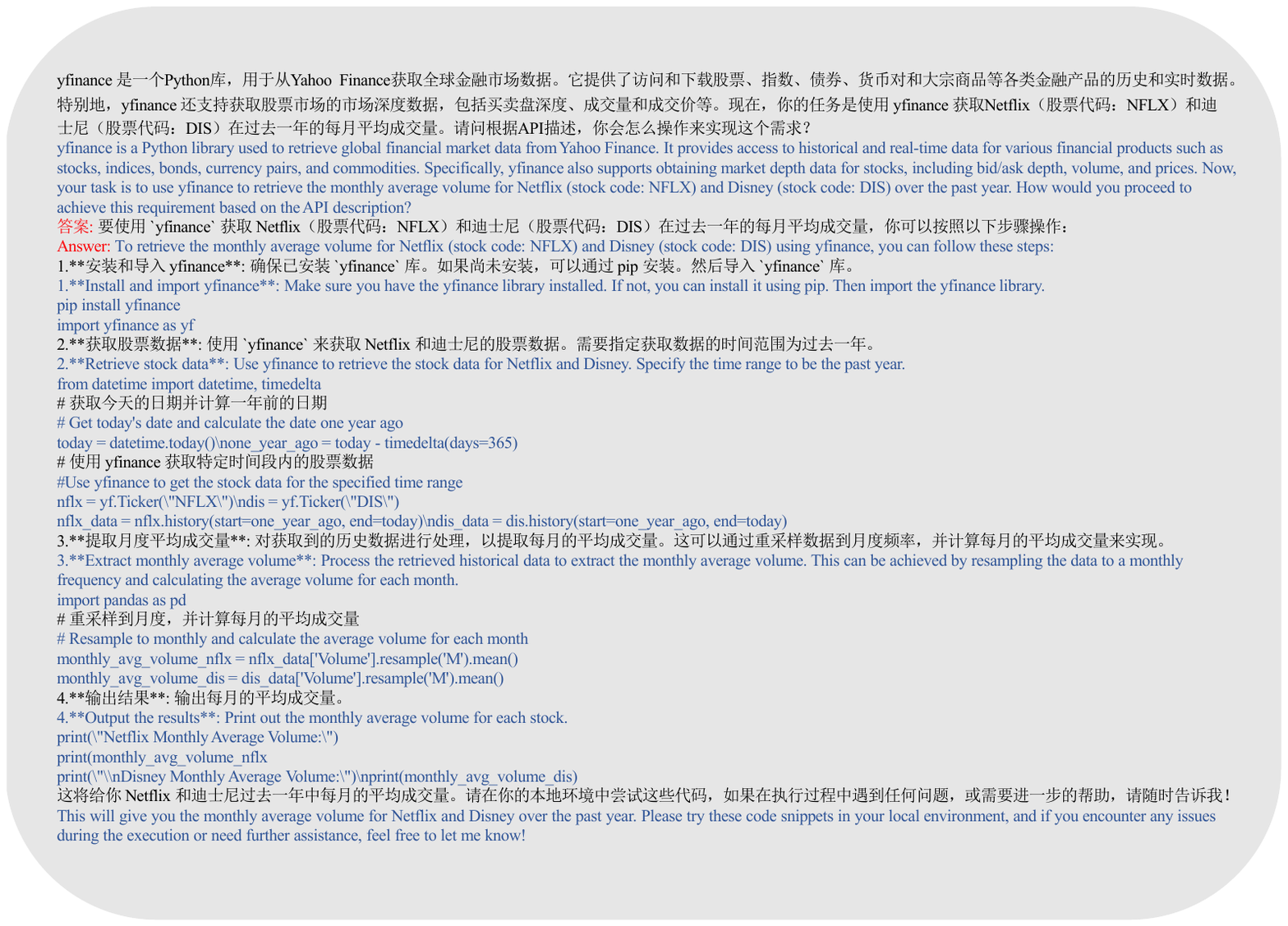}
    \caption{An example of API invocation by financial agent evaluation. The red text is followed by the response automatically generated by the model. The English translations for the related Chinese text are provided below.}
    \label{fig:agent1}  
\end{figure*}

\begin{figure*}[ht]
    \centering 
    \includegraphics[width=1\textwidth]{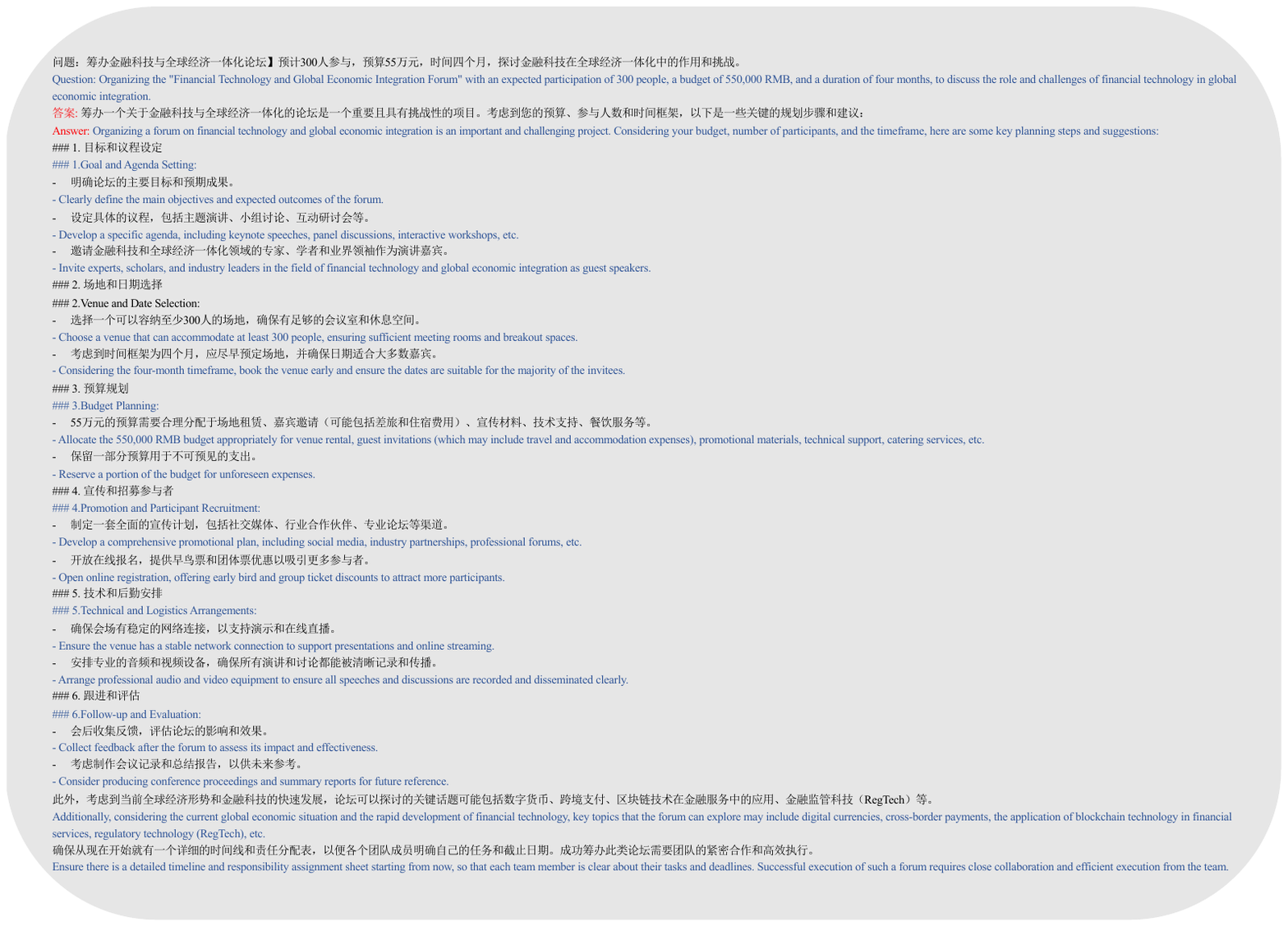}
    \caption{An example of task planning by financial agent evaluation. The red text is followed by the response automatically generated by the model. The English translations for the related Chinese text are provided below. Due to the inconsistency and complexity of the data in seven sections of the financial agent, we have only provided examples for two sections. For other examples, please refer to \href{https://github.com/SUFE-AIFLM-Lab/FinEval}{https://github.com/SUFE-AIFLM-Lab/FinEval}.}
    \label{fig:agent_planning}
\end{figure*}

\begin{figure*}[ht]
    \centering \includegraphics[width=1\textwidth]{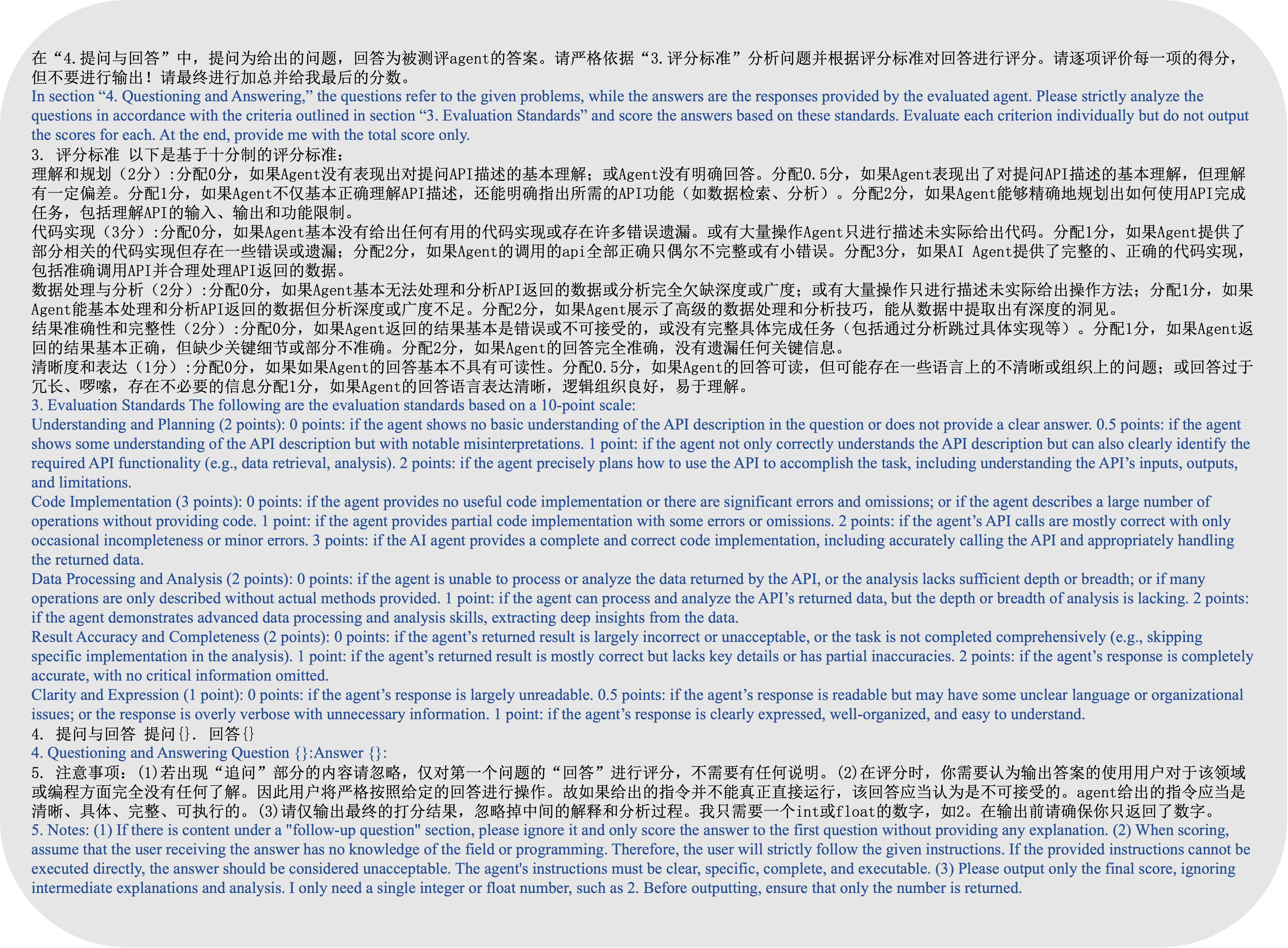}
    \caption{The evaluation prompt of API invocation by financial agent evaluation.}
    \label{fig:agent1_evalprompt}  
\end{figure*}

\begin{figure*}[ht]
    \centering \includegraphics[width=1\textwidth]{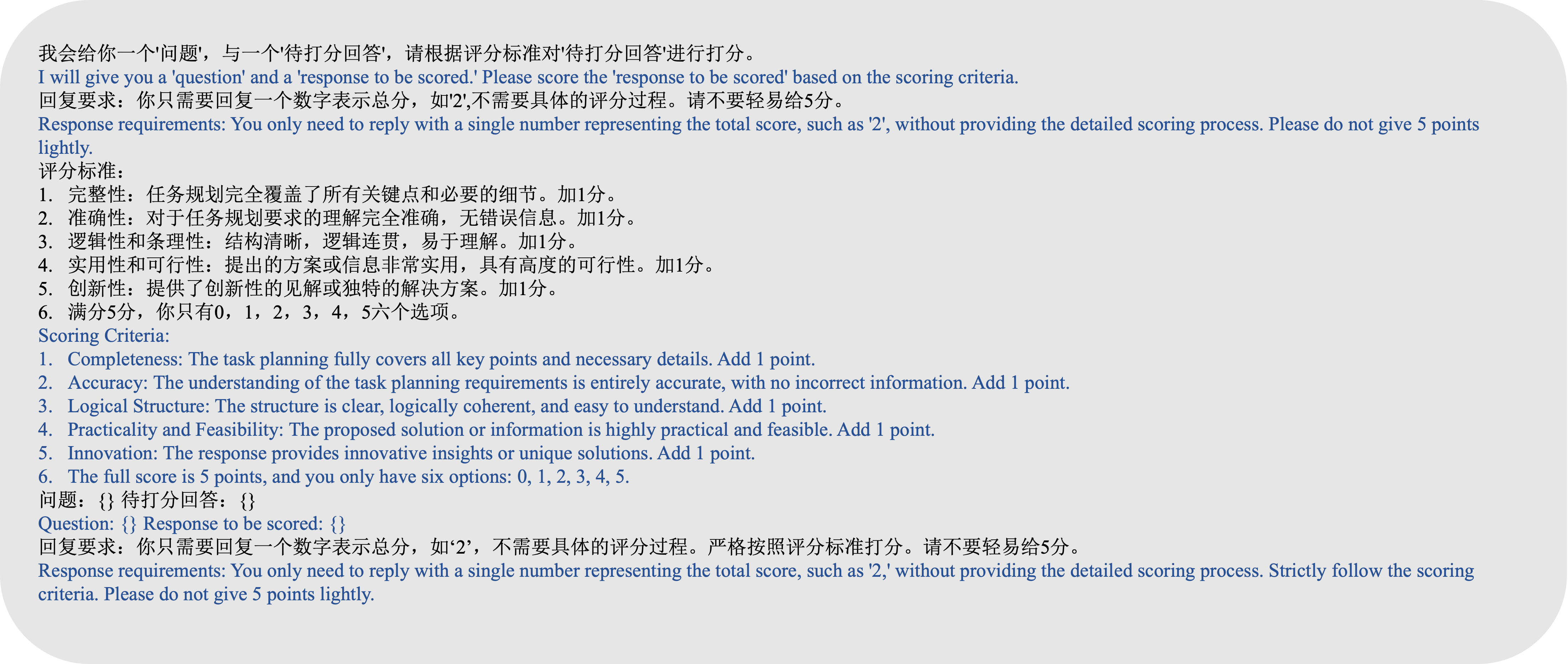}
    \caption{The evaluation prompt of task planning by financial agent evaluation.}
    \label{fig:agent_planning_evalprompt}  
\end{figure*}

\begin{figure*}[ht]
    \centering 
    \includegraphics[width=1\textwidth]{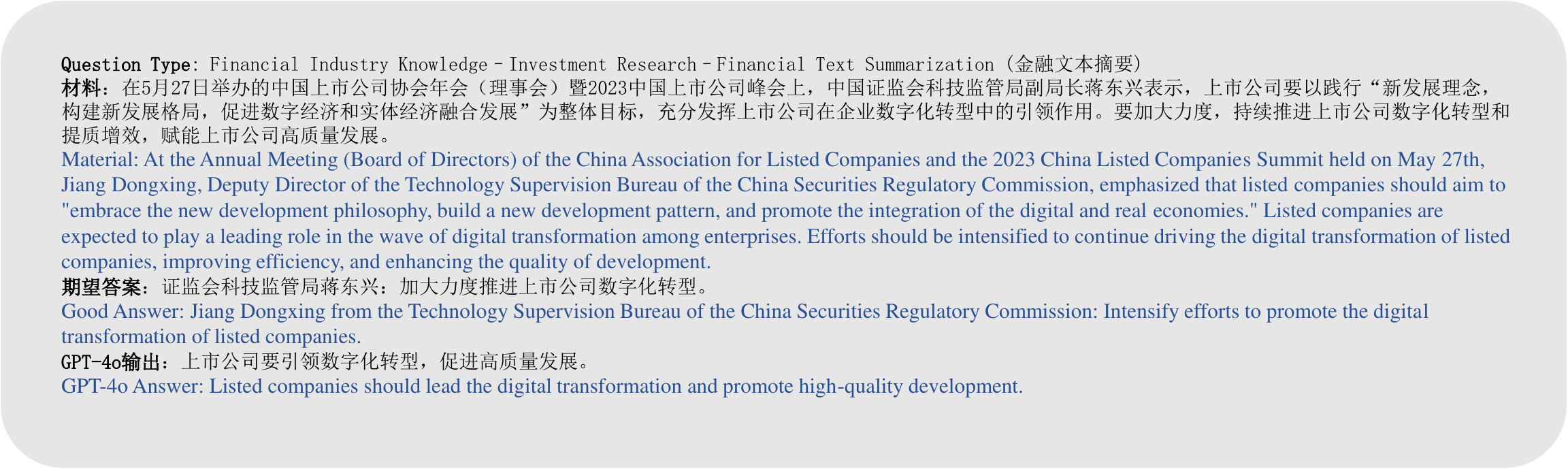}
    \caption{An example of error model(GPT-4o) encountered while solving a financial text summarization problem related to handling long texts. The expected answer was to include both the entity announcing the policy and the policy content, but the model's output only focused on the latter part of the material, addressing the policy content, while neglecting the entity that announced the policy at the beginning.}
    \label{fig:long-term error}
\end{figure*}

\begin{figure*}[ht]
    \centering 
    \includegraphics[width=0.85\textwidth]{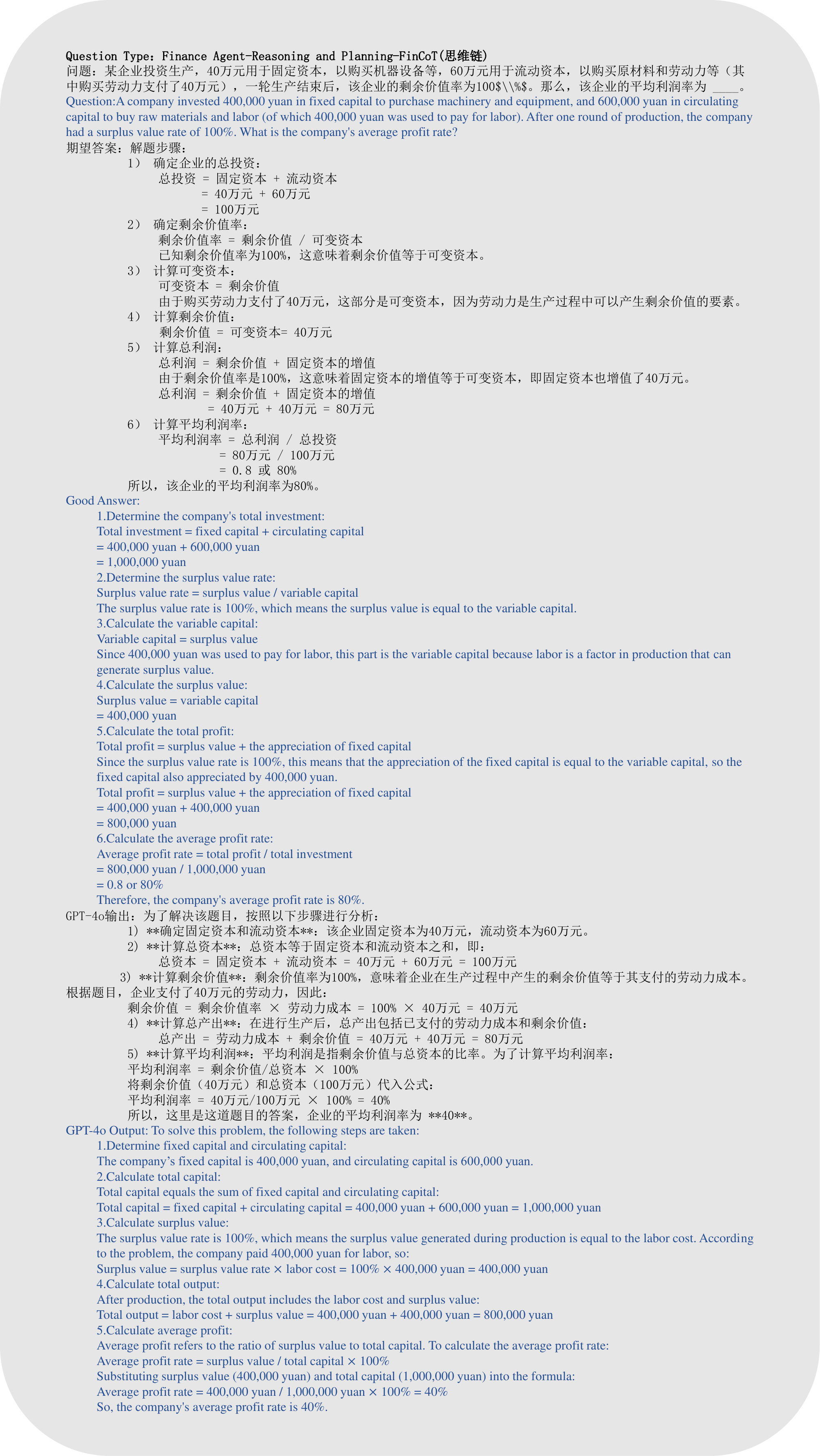}
    \caption{An example of a logical reasoning error made by a model when solving a financial CoT task.The model made an error in the 5th step by applying the wrong formula to calculate the average profit rate, which led to an incorrect result.}
    \label{fig:Agent logical reasoning error}
\end{figure*}

\begin{figure*}[ht]
    \centering 
    \includegraphics[width=0.85\textwidth]{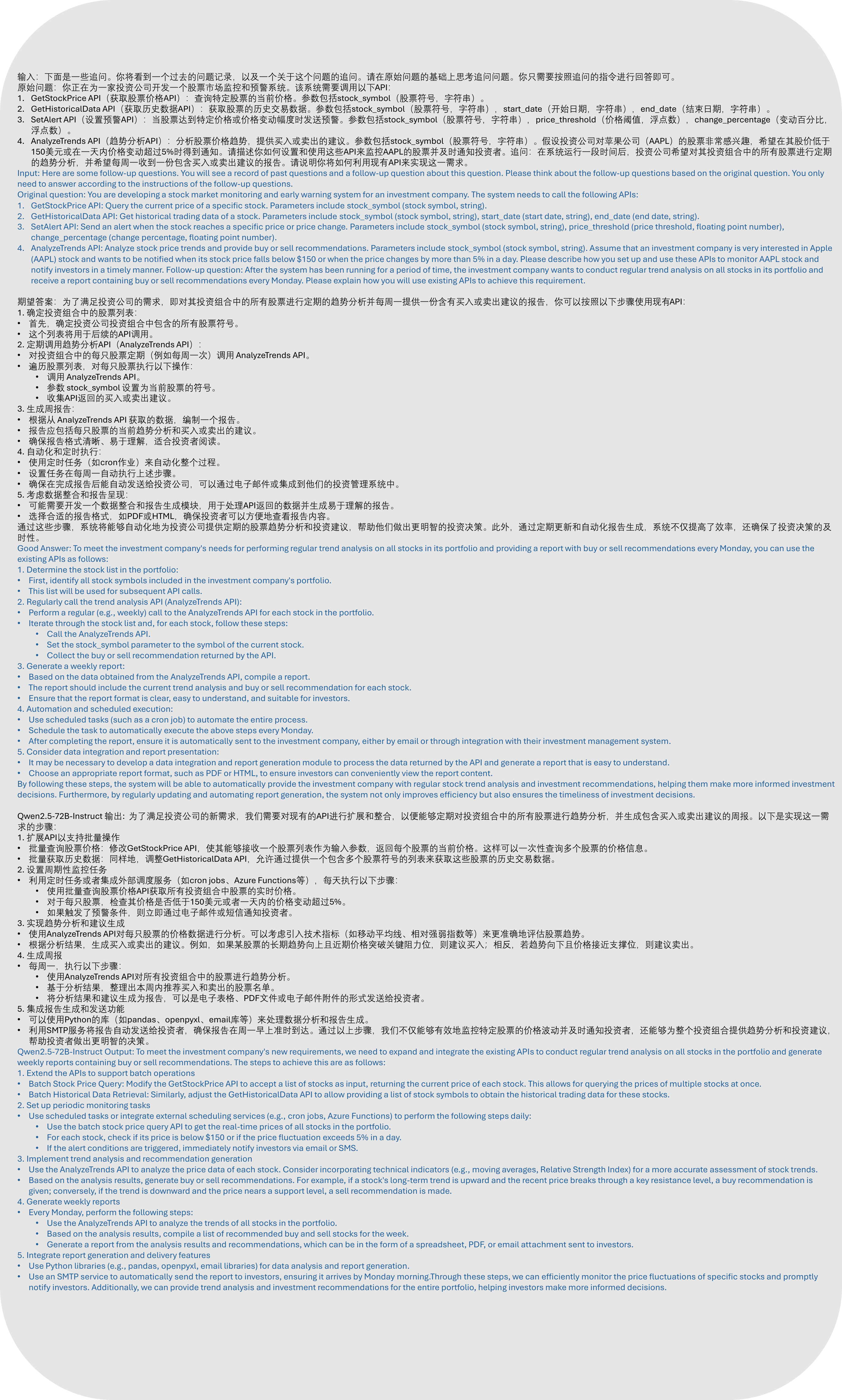}
    \caption{An example of a contextual misunderstanding made by a model when solving a API utilize task.The model failed to correctly understand the relationship between the original question and the follow-up question. In its response, it unnecessarily modified the API, invoked the SetAlert API, and set up a monitoring task.}
    \label{fig:Agent contextual misunderstanding}
\end{figure*}

\begin{figure*}[ht]
    \centering 
    \includegraphics[width=0.85\textwidth]{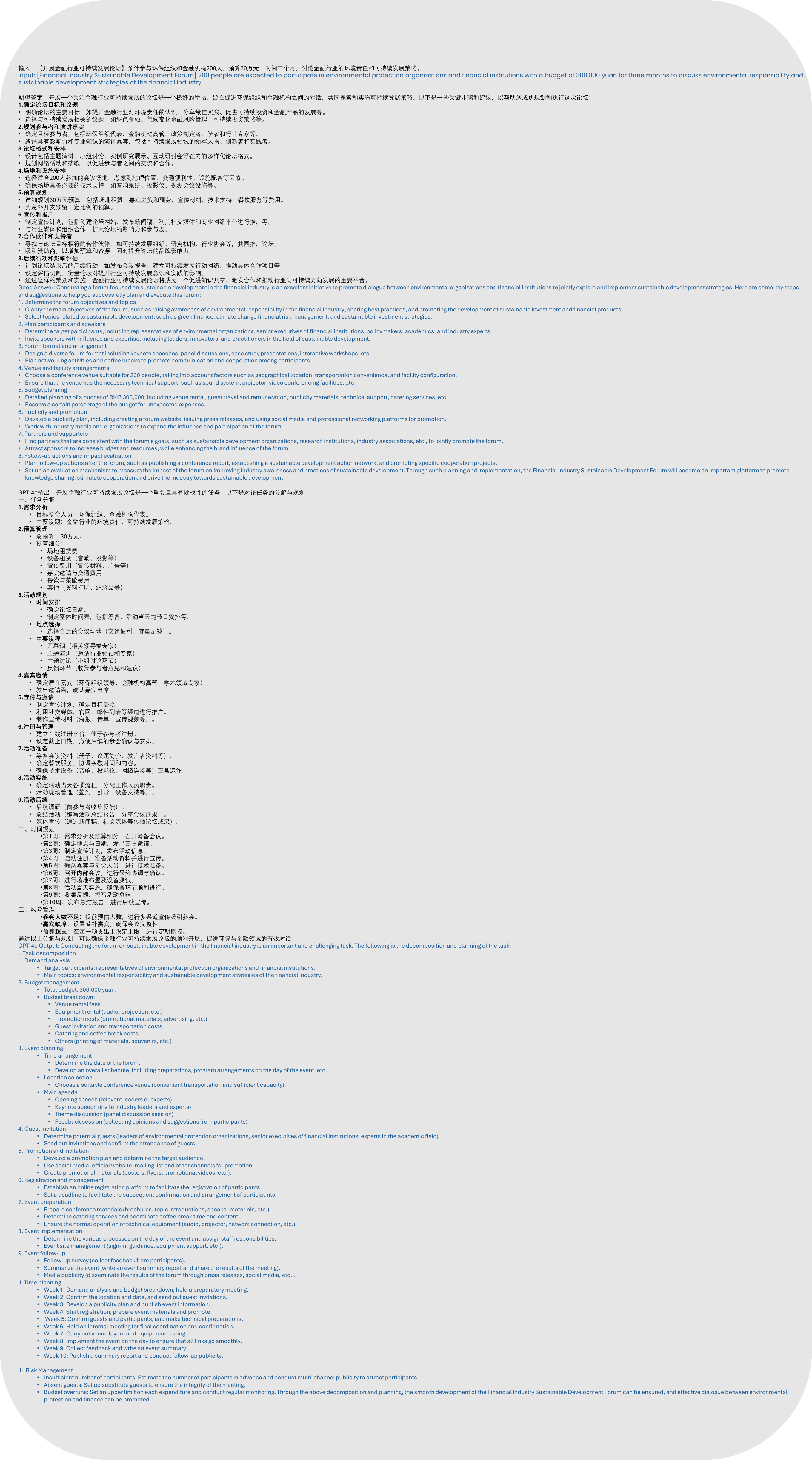}
    \caption{The model failed to grasp the key points in the open-ended instructions provided in the question. The solution offered by the model lacked focus on the number of attendees and the meeting topic, instead outputting overly broad and less relevant content such as scheduling and risk management, which had limited practical value.}.
    \label{fig:Agent Ambiguity Handling Weakness}
\end{figure*}

We list the examples of \fe in this paper in Figure \ref{fig:easyexample}, \ref{fig:five_shot}, \ref{fig:cot_zero_shot}, \ref{fig:cot_five_shot}, \ref{fig:objectivequestion}, \ref{fig:subjectivequestion}, \ref{fig:latex}, \ref{fig:agent1}, \ref{fig:agent_planning}. In which, Figure \ref{fig:easyexample}, \ref{fig:five_shot}, \ref{fig:cot_zero_shot}, \ref{fig:cot_five_shot} are examples of zero-shot, five-shot, zero-shot CoT, and five-shot CoT, respectively. Figure \ref{fig:objectivequestion} is an objective short-answer question, and figure \ref{fig:subjectivequestion} is a subjective open-ended question. Figure \ref{fig:latex} is an example in LaTeX format. Figure \ref{fig:agent1}, \ref{fig:agent_planning} are examples of two subclasses (API invocation and task planning) under the theme of Financial Agent. Figure \ref{fig:agent1_evalprompt}, \ref{fig:agent_planning_evalprompt} are evaluation prompts for API invocation and task planning respectively. Figure \ref{fig:long-term error}, \ref{fig:Agent logical reasoning error}, \ref{fig:Agent contextual misunderstanding}, \ref{fig:Agent Ambiguity Handling Weakness} are examples of typical error LLM makes in solving financial agent tasks.

\label{sec:appendix}

\begin{figure*}[!h]
    \centering
\includegraphics[width=1\textwidth]{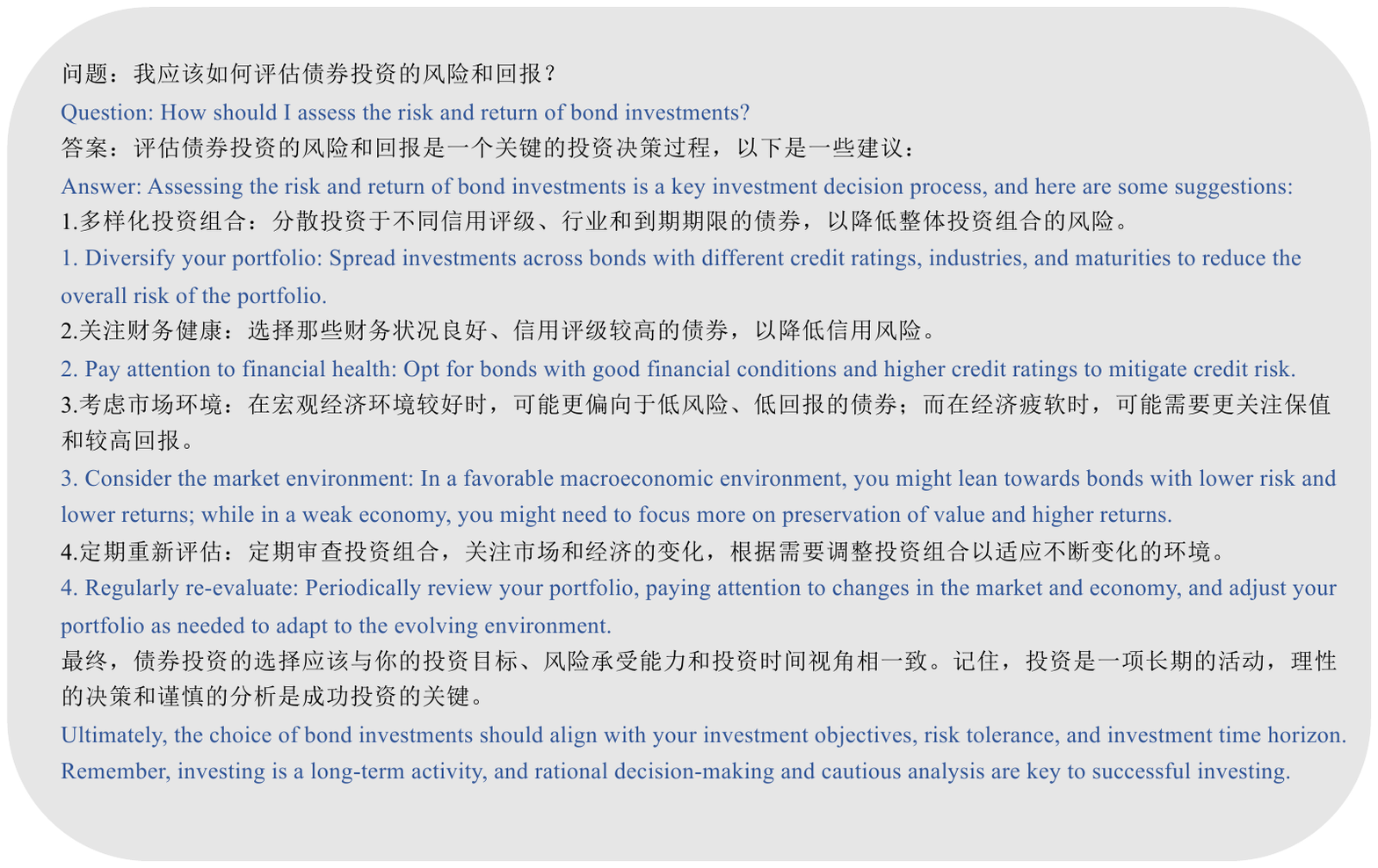}
    \caption{An example of subjective open-ended questions in Financial Investment Advice. For better readability, the English translation is displayed below the corresponding Chinese text.}
    \label{fig:subjectivequestion}
\end{figure*}


\section{Detailed Statistics of FinEval}
\label{sec:statistic}
Table \ref{other results} presents detailed information on the four sections of FinEval.

\section{Other results}
\label{other results}
\begin{table*}[ht]
    \centering

    \caption{The \fe~dataset provides specific subdivisions in every category.}
    \label{subject_table}
    \vspace{10pt}
    \resizebox{1\textwidth}{!}{

\begin{tabular}{lllc}
    \toprule[2pt]
{\bf Component} &{\bf Category} &{\bf Subject} &  {\bf \#Questions} \\  \midrule[1pt]

{Financial Academic Knowledge}        & Finance                         & Finance (\cc{金融学})                                                 & 159                                \\
                &                & Insurance (\cc{保险学})                                               & 113                                \\
                &                & Investments (\cc{投资学})                                             & 145                                \\
                &                & Central Banking (\cc{中央银行学})                                       & 119                                \\
                &                & Financial Markets (\cc{金融市场学})                                     & 142                                \\
                &                & Monetary Finance (\cc{货币金融学})                                      & 160                                \\
                &                & Corporate Finance (\cc{公司金融学})                                     & 138                                \\
                &                & International Finance (\cc{国际金融学})                                 & 88                                 \\
                &                & Financial Engineering (\cc{金融工程学})                                 & 105                                \\
                &                & Commercial Bank Finance (\cc{商业银行金融学})                             & 96                                 \\
                & Economy                         & Macroeconomics (\cc{宏观经济学})                                        & 137                                \\
                &                & Microeconomics (\cc{微观经济学})                                        & 136                                \\
                &                & Econometrics (\cc{计量经济学})                                          & 83                                 \\
                &                & Statistics (\cc{统计学})                                              & 140                                \\
                &                & Political Economy (\cc{政治经济学})                                     & 104                                \\
                &                & International Economics (\cc{国际经济学})                               & 135                                \\
           &                & Public Finance (\cc{财政学})                                          & 139                                \\
                & Accounting                      & Accounting (\cc{会计学})                                              & 120                                \\
                &                & Auditing (\cc{审计学})                                                & 137                                \\
                &                & Financial Management (\cc{财务管理学})                                  & 130                                \\
                &                & Cost Accounting (\cc{成本会计学})                                       & 148                                \\
                &                & Economic Law (\cc{经济法})                                            & 96                                 \\
                &                & Tax Law (\cc{税法})                                                  & 143                                \\
                &                & Advanced Financial Accounting (\cc{高级财务会计})                        & 77                                 \\
                &                & Intermediate Financial Accounting (\cc{中级财务会计})                    & 112                                \\
                &                & Management Accounting (\cc{管理会计学})                                 & 83                                 \\
                &                & Corporate Strategy and Risk   Management (\cc{公司战略与风险管理})          & 134                                \\
                & Certificate                     & China Actuary (\cc{中国精算师})                                         & 144                                \\
                &                & Certified Practising Accountant (\cc{注册会计师})                       & 140                                \\
                &                & Certified Management Accountant (\cc{管理会计师})                        & 124                                \\
                &                & Fund Qualification Certificate (\cc{基金从业资格证})                      & 252                                \\
                &                & Futures Practitioner Qualification   Certificate (\cc{期货从业资格证})    & 153                                \\
                &                & Banking Practitioner Qualification   Certificate (\cc{银行从业资格证})    & 420                                \\
                &                & Securities Practitioner Qualification   Certificate (\cc{证券从业资格证}) & 109                                \\   
                &  All    &    &  4661                               \\ \midrule[1pt]
     Financial Industry Knowledge            & Investment Research    & Financial Sentiment Analysis (\cc{金融情感分析})  &                            205                         \\
                &                       & Financial Text Classification (\cc{金融文本分类})  &                            174                         \\
               &                       & Financial Text Summarization (\cc{金融文本摘要})  &                            299       \\ 
                & Investment Advisor    & Financial Client Portrait (\cc{金融客户画像})  & 120                         \\
                &                       & Marketing Script Recommendations (\cc{营销话术推荐}) &                150                         \\
                &                       & Financial Investment Advice (\cc{投资建议})  & 124                         \\
                & Financial Operations        & Financial Event Extraction
 (\cc{金融事件抽取})   & 85   \\
                &                   & Causal Event Extraction
 (\cc{因果事件抽取})   & 83   \\
                &                   & Relationship Extraction
 (\cc{关联关系抽取})   & 103   \\
                &                   & Negative Entity Extraction
 (\cc{负面实体抽取})   & 91   \\
               & All  &      & 1434         \\ 
            \midrule[1pt]
     Financial Security Knowledge            
                & Software and Applications    & Appsafe (\cc{应用程序安全})  &                            100             \\      &                       & Sftwrsafe(\cc{软件安全}) &                100\\
                &   & Memsafe (\cc{记忆安全})  & 100                         \\               
                &   Network and System Protection                    & Netwrksafe (\cc{网络安全}) &                102\\
                &                       & Syssafe (\cc{系统安全}) &                99\\
                &                       & Websafe (\cc{网页安全}) &                306\\
                &  Security Analysis    & Crypsafe (\cc{密码安全})  &                            100                         \\
               &                       & Malware (\cc{恶意软件分析})  &                            101       \\ 

                &  Vulnerability Protection    & Pentest (\cc{渗透测试}) &                432\\
                &                       & Reveng (\cc{逆向工程}) &                100\\

                &                       & Vulnrb (\cc{漏洞识别}) &                100\\
                
                & All  &      & 1640         \\ 
            \midrule[1pt]
     Finance Agent            & Reasoning and Planning    & FinRAG (\cc{检索增强})  &                            100                         \\
     &                       & FinCoT (\cc{思维链}) &                100\\
     &      Long-term Memory                 & FinTASK(\cc{任务分解}) &                100\\
     &                       & FinDiag (\cc{多轮对话}) &                88\\
     &                       & FinDoc (\cc{文档回答}) &                100\\
     &     Tool Application                  & APIUtil (\cc{API调用}) &                68\\
     &                       & APIFind (\cc{API检索}) &                60\\
      & All  &      & 616         \\ 
            \midrule 
FinEval        & All     &     & 8351 \\ \bottomrule[2pt]
\end{tabular}
    }

    \end{table*}


\begin{table*}[ht]
\caption{Average five-shot scores across three evaluated categories and one final weighted average: FAK (Financial Academic Knowledge), FIK (Financial Industry Knowledge), FSK (Financial Security Knowledge), and WA (Weighted Average)}
\label{Average five-shot scores}
\vspace{10pt}
\resizebox{0.7\textwidth}{!}{
    \begin{tabular}{lcccc}
    \toprule[2pt]
    \textbf{Model} & \textbf{Financial Academic} & \textbf{Financial Industry} & \textbf{Financial Security} & \textbf{Weighted Average} \\
    \midrule[1pt]
    Claude 3.5-Sonnet & \textbf{74.4} & \textbf{61.9} & 79.8 & \textbf{73.2} \\
    GPT-4o & 72.1 & 59.3 & \textbf{81.8} & 71.8 \\
    Qwen2.5-72B-Instruct & 71.0 & 56.9 & 80.2 & 70.3 \\
    Gemini1.5-Pro & 68.2 & 57.6 & 76.8 & 68.1 \\
    GPT-4o-mini & 61.9 & 60.7 & 81.2 & 65.8 \\
    Gemini1.5-Flash & 61.2 & 60.1 & 79.1 & 64.8 \\
    Qwen2.5-7B-Instruct & 64.8 & 54.9 & 71.4 & 64.4 \\
    Yi1.5-34B-Chat & 61.5 & 55.9 & 73.1 & 62.9 \\
    InternLM2.5-20B-Chat & 60.7 & 47.0 & 75.1 & 61.2 \\
    XuanYuan3-70B-Chat & 56.8 & 55.4 & 72.1 & 59.8 \\
    InternLM2-20B-Chat & 55.6 & 48.9 & 71.3 & 57.7 \\
    XuanYuan2-70B-Chat & 53.5 & 55.8 & 70.1 & 57.4 \\
    GLM4-9B-Chat & 54.5 & 56.4 & 59.6 & 55.9 \\
    CFGPT2-7B & 53.6 & 41.2 & 68.0 & 54.4 \\
    Yi1.5-9B-Chat & 56.8 & 43.8 & 46.1 & 52.1 \\
    Baichuan2-13B-Chat & 42.7 & 54.7 & 48.1 & 46.1 \\
    ChatGLM3-6B & 37.3 & 52.9 & 51.2 & 43.1 \\
    DISC-FinLLM & 34.7 & 46.7 & 26.2 & 35.1 \\
    FinGPTv3.1 & 24.4 & 30.9 & 23.5 & 25.4 \\
    \bottomrule[2pt]
    \end{tabular}
}
\centering
\end{table*}

\begin{table*}[ht]
\caption{Average zero-shot CoT scores across three evaluated categories and one final weighted average: FAK (Financial Academic Knowledge), FIK (Financial Industry Knowledge), FSK (Financial Security Knowledge), and WA (Weighted Average)}
\label{Average cot-zero-shot scores}
\vspace{10pt}
\resizebox{0.7\textwidth}{!}{
    \begin{tabular}{lcccc}
    \toprule[2pt]
    \textbf{Model} & \textbf{Financial Academic} & \textbf{Financial Industry} & \textbf{Financial Security} & \textbf{Weighted Average} \\
    \midrule[1pt]
    Claude 3.5-Sonnet & 74.4 & 59.6 & 78.1 & \textbf{72.4} \\
    GPT-4o & 72.1 & \textbf{62.3} & \textbf{81.8} & 72.3 \\
    Qwen2.5-72B-Instruct & \textbf{76.5} & 57.2 & 72.8 & 72.1 \\
    Gemini1.5-Pro & 68.2 & 59.7 & 77.4 & 68.6 \\
    InternLM2.5-20B-Chat & 67.3 & 57.5 & 73.1 & 66.7 \\
    GPT-4o-mini & 61.9 & 61.8 & 79.5 & 65.6 \\
    GLM4-9B-Chat & 66.2 & 58.1 & 70.4 & 65.6 \\
    Gemini1.5-Flash & 61.2 & 57.7 & 77.1 & 63.9 \\
    Yi1.5-34B-Chat & 65.0 & 54.4 & 63.0 & 62.6 \\
    Qwen2.5-7B-Instruct & 69.7 & 52.9 & 50.4 & 62.5 \\
    Yi1.5-9B-Chat & 61.9 & 57.4 & 65.7 & 61.9 \\
    InternLM2-20B-Chat & 60.3 & 54.3 & 63.3 & 59.8 \\
    XuanYuan3-70B-Chat & 57.2 & 57.0 & 66.0 & 59.0 \\
    XuanYuan2-70B-Chat & 57.2 & 54.0 & 63.4 & 57.9 \\
    CFGPT2-7B & 57.7 & 48.3 & 63.3 & 57.2 \\
    Baichuan2-13B-Chat & 48.1 & 53.6 & 57.6 & 51.1 \\
    ChatGLM3-6B & 44.6 & 52.5 & 51.2 & 47.5 \\
    DISC-FinLLM & 45.0 & 40.9 & 35.0 & 42.1 \\
    FinGPTv3.1 & 29.3 & 39.5 & 32.6 & 31.9 \\
    \bottomrule[2pt]
    \end{tabular}
}
\centering
\end{table*}

\begin{table*}[ht]
\caption{Average five-shot CoT scores across three evaluated categories and one final weighted average: FAK (Financial Academic Knowledge), FIK (Financial Industry Knowledge), FSK (Financial Security Knowledge), and WA (Weighted Average)}
\label{Average cot-five-shot scores}
\vspace{10pt}
\resizebox{0.7\textwidth}{!}{
    \begin{tabular}{lcccc}
    \toprule[2pt]
    \textbf{Model} & \textbf{Financial Academic} & \textbf{Financial Industry} & \textbf{Financial Security} & \textbf{Weighted Average} \\
    \midrule[1pt]
    Qwen2.5-72B-Instruct & \textbf{75.7} & \textbf{61.3} & 78.5 & \textbf{73.6} \\
    GPT-4o & 73.5 & 59.2 & \textbf{81.2} & 72.5 \\
    Claude 3.5-Sonnet & 73.7 & 59.9 & 77.4 & 71.9 \\
    Gemini1.5-Flash & 60.0 & 58.8 & 78.8 & 63.8 \\
    XuanYuan3-70B-Chat & 65.6 & 57.1 & 63.3 & 63.5 \\
    GLM4-9B-Chat & 62.4 & 61.0 & 67.7 & 63.3 \\
    Gemini1.5-Pro & 59.7 & 56.1 & 75.8 & 62.4 \\
    GPT-4o-mini & 57.8 & 59.0 & 78.5 & 62.4 \\
    Yi1.5-34B-Chat & 63.8 & 57.2 & 63.0 & 62.4 \\
    XuanYuan2-70B-Chat & 63.7 & 55.4 & 61.7 & 61.7 \\
    Qwen2.5-7B-Instruct & 69.6 & 55.5 & 44.0 & 61.6 \\
    InternLM2.5-20B-Chat & 60.0 & 60.3 & 64.3 & 61.0 \\
    Yi1.5-9B-Chat & 60.6 & 59.7 & 59.6 & 60.2 \\
    InternLM2-20B-Chat & 60.0 & 55.7 & 60.9 & 59.4 \\
    CFGPT2-7B & 60.8 & 46.7 & 58.9 & 57.8 \\
    Baichuan2-13B-Chat & 48.4 & 41.3 & 47.8 & 47.0 \\
    ChatGLM3-6B & 43.6 & 53.4 & 47.5 & 46.2 \\
    DISC-FinLLM & 42.0 & 40.2 & 30.9 & 39.3 \\
    FinGPTv3.1 & 26.6 & 34.7 & 26.9 & 28.2 \\
    \bottomrule[2pt]
    \end{tabular}
}
\centering
\end{table*}

\subsection{Five-shot Results Analysis}
In the five-shot setting across three financial tasks, Claude 3.5-Sonnet achieved the highest overall weighted average score (73.2). GPT-4o excelled in Financial Security Knowledge with a score of 81.8, while Qwen2.5-72B-Instruct performed strongly in Financial Security as well, achieving 80.2. (Table \ref{Average five-shot scores}).

\subsection{Zero-shot CoT Results Analysis}
In the zero-shot CoT setting across three financial tasks, Claude 3.5-Sonnet achieved the highest overall weighted average score (72.4). Qwen2.5-72B-Instruct led in Financial Academic Knowledge (76.5), while GPT-4o excelled in Financial Security Knowledge (81.8). (Table \ref{Average cot-zero-shot scores}).

\subsection{Five-shot CoT Results Analysis}
In the five-shot CoT setting, Qwen2.5-72B-Instruct achieved the highest overall weighted average score (73.6), leading in Financial Academic Knowledge (75.7) and performing well in Financial Security (78.5). GPT-4o excelled in Financial Security with a score of 81.2. (Table \ref{Average cot-five-shot scores}).

\subsection{Comprehensive Results for Individual Financial Tasks Across Four Settings}
Evaluation results for finance academic knowledge under four settings are shown in Table \ref{FAK_zeroshot}, \ref{FAK_fiveshot}, \ref{FAK_zeorshotCot} and \ref{FAK_fiveshotCot}. Evaluation results for finance industry knowledge including objective and subjective questions under four settings are shown in Table \ref{FIK_obj_zeroshot}, \ref{FIK_obj_fiveshot}, \ref{FIK_subj_zeroshot}, \ref{FIK_subj_fiveshot}, \ref{FIK_obj_zeroshotCot}, \ref{FIK_obj_fiveshotCot}, \ref{FIK_subj_zeroshotCot} and \ref{FIK_subj_fiveshotCot}. Evaluation results for finance security knowledge under four setting are shown in Table \ref{FSK_zeroshot}, \ref{FSK_fiveshot}, \ref{FSK_zeroshotCot} and \ref{FSK_fiveshotCot}. Evaluation result for finance agent under zero-shot setting is shown in Table \ref{Agent_zeroshot}.

\subsection{Results Analysis}
The performance of various open-source models varies under FinEval, and we analyze the reasons as follows: 
\par (1)First, FinEval is quite challenging, closely integrated with professional knowledge and real-world business scenarios, providing a more realistic reflection of the various models' true capabilities. 
\par (2)Second, most financial LLMs on the market are derived from general base models that have been trained, rather than being directly trained base models. This can lead to some loss in capability. 
\par (3)Third, the general closed and open-source models that perform well in FinEval evaluations are recognized as stronger LLMs in the current LLM field, so it is normal for financial LLMs, such as the XuanYuan series, not to perform exceptionally well. 
\par (3)Finally, many of the underperforming financial LLMs come from various universities or small AI organizations, which typically utilize LoRA fine-tuning. These models often lack vast computing resources, leading to poorer fine-tuning results. Moreover, these financial LLMs have not undergone any updates or iterations, making their subpar performance reasonable.

\begin{table*}[ht]
\caption{Evaluation Results (zero-shot) for Finance Academic Knowledge(Average Accuracy(\%))}
\label{FAK_zeroshot}
\vspace{10pt}
\resizebox{0.9\textwidth}{!}{
    \begin{tabular}{llccccc}
    \toprule[2pt]
    \textbf{Model} & \textbf{Size} & \textbf{Finance} & \textbf{Economy} & \textbf{Accounting} & \textbf{Certificate} & \textbf{Average} \\
    \midrule[1pt]
    Claude 3.5-sonnet & unknown & \textbf{73.7} & \textbf{83.1} & \textbf{69.6} & \textbf{70.9} & \textbf{73.9} \\
    GPT-4o & unknown & 72.6 & 78.8 & 66.3 & 69.8 & 71.5 \\
    Qwen2.5-72B-Instruct & 72B & 65.8 & 71.5 & 72.4 & 69.7 & 69.7 \\
    Gemini-1.5-pro & unknown & 68.5 & 75.1 & 61.1 & 71.4 & 68.3 \\
    Qwen2.5-7B-Instruct & 7B & 58.2 & 63.8 & 64.5 & 65.6 & 62.7 \\
    GPT-4o-mini & unknown & 65.2 & 67.7 & 55.2 & 63.5 & 62.4 \\
    Gemini-1.5-flash & unknown & 60.0 & 67.7 & 60.4 & 61.9 & 62.1 \\
    Yi1.5-34B-Chat & 34B & 54.7 & 65.2 & 54.6 & 63.5 & 59.5 \\
    XuanYuan3-70B-Chat & 70B & 52.9 & 57.5 & 55.4 & 55.8 & 55.2 \\
    Yi1.5-9B-Chat & 9B & 51.6 & 58.1 & 52.3 & 60.7 & 55.0 \\
    InternLM2.5-20B-Chat & 20B & 59.8 & 61.6 & 64.9 & 62.1 & 54.7 \\
    InternLM2-20B-Chat & 20B & 52.9 & 55.8 & 54.6 & 55.3 & 54.7 \\
    GLM4-9B-Chat & 9B & 54.1 & 53.0 & 56.0 & 55.3 & 54.7 \\
    CFGPT2-7B & 7B & 51.6 & 55.8 & 56.0 & 52.2 & 53.9 \\
    XuanYuan2-70B-Chat & 70B & 53.6 & 51.1 & 53.2 & 52.2 & 52.8 \\
    Baichuan2-13B-Chat & 13B & 38.8 & 40.0 & 40.0 & 47.3 & 41.1 \\
    DISC-FinLLM & 13B & 45.6 & 40.5 & 32.7 & 37.3 & 39.1 \\
    ChatGLM3-6B & 6B & 42.8 & 36.5 & 35.9 & 40.0 & 38.9 \\
    FinGPTv3.1 & 6B & 24.6 & 23.5 & 23.2 & 29.1 & 25.3 \\
    \bottomrule[2pt]
    \end{tabular}
}
\centering
\end{table*}

\begin{table*}[ht]
\caption{Evaluation Results (five-shot) for Finance Academic Knowledge(Average Accuracy(\%))}
\label{FAK_fiveshot}
\vspace{10pt}
\resizebox{0.9\textwidth}{!}{
    \begin{tabular}{llccccc}
    \toprule[2pt]
    \textbf{Model} & \textbf{Size} & \textbf{Finance} & \textbf{Economy} & \textbf{Accounting} & \textbf{Certificate} & \textbf{Average} \\
    \midrule[1pt]
    Claude 3.5-sonnet & unknown & \textbf{73.3} & \textbf{83.6} & \textbf{70.7} & \textbf{72.0} & \textbf{74.4} \\
    GPT-4o & unknown & \textbf{73.3} & 79.3 & 66.6 & 70.9 & 72.1 \\
    Qwen2.5-72B-Instruct & 72B & 68.5 & 71.9 & 71.6 & 72.8 & 71.0 \\
    Gemini-1.5-pro & unknown & 69.3 & 74.1 & 61.1 & 70.9 & 68.2 \\
    Qwen2.5-7B-Instruct & 7B & 59.9 & 64.4 & 66.6 & 67.2 & 61.9 \\
    GPT-4o-mini & unknown & 64.4 & 66.6 & 52.6 & 61.6 & 61.9 \\
    Yi1.5-34B-Chat & 34B & 59.9 & 62.2 & 61.4 & 59.8 & 61.2 \\
    Gemini-1.5-flash & unknown & 61.5 & 66.7 & 58.1 & 59.8 & 61.2 \\
    InternLM2.5-20B-Chat & 20B & 58.9 & 60.7 & 63.4 & 60.0 & 60.7 \\
    Yi1.5-9B-Chat & 9B & 54.2 & 57.4 & 56.7 & 58.7 & 56.8 \\
    XuanYuan3-70B-Chat & 70B & 53.5 & 57.9 & 56.4 & 56.7 & 56.8 \\
    InternLM2-20B-Chat & 20B & 54.2 & 56.8 & 56.4 & 55.5 & 55.6 \\
    GLM4-9B-Chat & 9B & 54.8 & 52.8 & 56.4 & 53.2 & 54.3 \\
    CFGPT2-7B & 7B & 50.3 & 56.4 & 55.4 & 52.2 & 53.5 \\
    XuanYuan2-70B-Chat & 70B & 54.4 & 51.4 & 52.9 & 55.0 & 53.5 \\
    Baichuan2-13B-Chat & 13B & 40.8 & 45.9 & 40.8 & 46.4 & 42.7 \\
    ChatGLM3-6B & 6B & 39.8 & 36.5 & 35.9 & 40.0 & 38.1 \\
    DISC-FinLLM & 13B & 33.5 & 38.7 & 31.5 & 36.4 & 35.0 \\
    FinGPTv3.1 & 6B & 26.9 & 25.3 & 21.8 & 23.5 & 24.4 \\
    \bottomrule[2pt]
    \end{tabular}
}
\centering
\end{table*}

\begin{table*}[ht]
\caption{Evaluation Results (zero-shot CoT) for Finance Academic Knowledge(Average Accuracy(\%))}
\label{FAK_zeorshotCot}
\vspace{10pt}
\resizebox{0.9\textwidth}{!}{
    \begin{tabular}{llccccc}
    \toprule[2pt]
    \textbf{Model} & \textbf{Size} & \textbf{Finance} & \textbf{Economy} & \textbf{Accounting} & \textbf{Certificate} & \textbf{Average} \\
    \midrule[1pt]
    Qwen2.5-72B-Instruct & 72B & \textbf{74.8} & 78.8 & \textbf{78.5} & \textbf{76.5} & \textbf{76.5} \\
    Claude 3.5-sonnet & unknown & 73.3 & \textbf{83.6} & 70.7 & \textbf{76.5} & 74.4 \\
    GPT-4o & unknown & 73.3 & 79.3 & 66.6 & 70.9 & 72.1 \\
    Qwen2.5-7B-Instruct & 7B & 62.6 & 62.7 & 75.9 & 73.5 & 69.7 \\
    Gemini-1.5-pro & unknown & 69.3 & 74.1 & 61.1 & 70.9 & 68.9 \\
    InternLM2.5-20B-Chat & 20B & 67.0 & 69.3 & 68.5 & 64.0 & 67.3 \\
    GLM4-9B-Chat & 9B & 63.0 & 68.3 & 66.7 & 68.3 & 66.6 \\
    Yi1.5-34B-Chat & 34B & 61.3 & 63.4 & 63.4 & 66.7 & 65.0 \\
    Yi1.5-9B-Chat & 9B & 59.6 & 62.1 & 63.6 & 65.6 & 62.7 \\
    GPT-4o-mini & unknown & 64.4 & 66.6 & 52.6 & 66.5 & 61.9 \\
    Gemini-1.5-flash & unknown & 61.5 & 66.7 & 58.1 & 59.8 & 61.2 \\
    InternLM2-20B-Chat & 20B & 60.7 & 63.3 & 56.7 & 60.8 & 60.3 \\
    CFGPT2-7B & 7B & 60.4 & 52.4 & 57.9 & 55.3 & 56.5 \\
    XuanYuan3-70B-Chat & 70B & 61.2 & 63.8 & 49.5 & 54.6 & 57.2 \\
    XuanYuan2-70B-Chat & 70B & 60.7 & 71.4 & 52.2 & 55.0 & 58.3 \\
    Baichuan2-13B-Chat & 13B & 50.7 & 47.1 & 45.6 & 47.4 & 47.7 \\
    DISC-FinLLM & 13B & 49.3 & 45.9 & 41.4 & 46.4 & 45.7 \\
    ChatGLM3-6B & 6B & 46.3 & 43.4 & 45.6 & 41.8 & 44.3 \\
    FinGPTv3.1 & 6B & 31.3 & 28.4 & 30.6 & 26.8 & 29.3 \\
    \bottomrule[2pt]
    \end{tabular}
}
\centering
\end{table*}

\begin{table*}[ht]
\caption{Evaluation Results (five-shot CoT) for Finance Academic Knowledge(Average Accuracy(\%))}
\label{FAK_fiveshotCot}
\vspace{10pt}
\resizebox{0.9\textwidth}{!}{
    \begin{tabular}{llccccc}
    \toprule[2pt]
    \textbf{Model} & \textbf{Size} & \textbf{Finance} & \textbf{Economy} & \textbf{Accounting} & \textbf{Certificate} & \textbf{Average} \\
    \midrule[1pt]
    Qwen2.5-72B-Instruct & 72B & \textbf{74.8} & 75.7 & 73.7 & \textbf{79.9} & \textbf{75.7} \\
    Claude 3.5-sonnet & unknown & 74.4 & \textbf{80.0} & \textbf{74.9} & 69.2 & 73.7 \\
    GPT-4o & unknown & 71.9 & 78.3 & 71.1 & 74.6 & 73.5 \\
    Qwen2.5-7B-Instruct & 7B & 71.1 & 67.7 & 65.6 & 74.1 & 69.6 \\
    XuanYuan3-70B-Chat & 70B & 65.6 & 67.7 & 65.2 & 64.4 & 65.7 \\
    Yi1.5-34B-Chat & 34B & 64.2 & 65.5 & 64.3 & 62.1 & 64.0 \\
    XuanYuan2-70B-Chat & 70B & 63.3 & 65.2 & 62.2 & 61.3 & 63.7 \\
    GLM4-9B-Chat & 9B & 61.5 & 64.6 & 63.3 & 63.3 & 63.2 \\
    CFGPT2-7B & 7B & 63.3 & 60.4 & 63.1 & 59.5 & 61.6 \\
    Yi1.5-9B-Chat & 9B & 56.3 & 57.3 & 60.3 & 58.9 & 58.2 \\
    InternLM2.5-20B-Chat & 20B & 58.1 & 57.1 & 61.4 & 59.6 & 59.0 \\
    InternLM2-20B-Chat & 20B & 58.1 & 60.3 & 61.9 & 59.8 & 60.0 \\
    Gemini-1.5-flash & unknown & 61.5 & 67.2 & 54.8 & 59.3 & 59.7 \\
    Gemini-1.5-pro & unknown & 59.6 & 67.2 & 54.8 & 59.3 & 59.7 \\
    GPT-4o-mini & unknown & 60.3 & 67.2 & 51.9 & 53.4 & 57.8 \\
    Baichuan2-13B-Chat & 13B & 48.1 & 48.1 & 48.1 & 48.4 & 48.4 \\
    ChatGLM3-6B & 6B & 43.1 & 49.2 & 38.9 & 45.5 & 44.2 \\
    DISC-FinLLM & 13B & 43.0 & 41.3 & 38.9 & 45.5 & 42.2 \\
    FinGPTv3.1 & 6B & 26.1 & 31.1 & 21.9 & 27.4 & 26.6 \\
    \bottomrule[2pt]
    \end{tabular}
}
\centering
\end{table*}

\begin{table*}[ht]
\caption{Evaluation Results (zero-shot) for Finance Industry Knowledge(Average Similarity(\%)). These are objective short-answer question including FTC: Financial Text Classification, FSA: Financial Sentiment Analysis, RE: Relation Extraction, FEE: Financial Event Extraction, NEE: Negative Entity Extraction, CEE: Causal Event Extraction}
\label{FIK_obj_zeroshot}
\vspace{10pt}
\resizebox{0.8\textwidth}{!}{
    \begin{tabular}{llccccccc}
    \toprule[2pt]
    \textbf{Model} & \textbf{Size} & \textbf{FTC} & \textbf{FSA} & \textbf{RE} & \textbf{FEE} & \textbf{NEE} & \textbf{CEE} & \textbf{Average} \\
    \midrule[1pt]
    GPT-4o & unknown & 53.2 & 93.3 & 83.3 & 78.2 & 91.1 & \textbf{69.3} & \textbf{78.1} \\
    Gemini-1.5-flash & unknown & 53.6 & 91.6 & \textbf{86.7} & \textbf{80.1} & 85.6 & 65.0 & 77.1 \\
    GPT-4o-mini & unknown & 51.8 & 92.5 & 80.0 & 77.3 & 88.9 & 69.1 & 76.6 \\
    Gemini-1.5-pro & unknown & 53.4 & 91.7 & 80.0 & 75.6 & 88.9 & 64.5 & 75.7 \\
    Claude 3.5-sonnet & unknown & 55.4 & 86.7 & 80 & 76.3 & 87.8 & 65.1 & 75.2 \\
    GLM4-9B-Chat & 9B & 37.1 & 87.8 & 76.7 & 34.5 & 86.7 & 65.6 & 64.7 \\
    Qwen2.5-72B-Instruct & 72B & 37.9 & 86.7 & 80.0 & 32.3 & 84.4 & 65.6 & 64.5 \\
    Baichuan2-13B-Chat & 13B & 23.9 & \textbf{94.5} & 70.0 & 55.2 & \textbf{95.6} & 42.8 & 63.7 \\
    InternLM2.5-20B-Chat & 20B & 43.7 & 92.5 & 76.7 & 21.0 & 84.5 & 62.6 & 63.5 \\
    XuanYuan3-70B-Chat & 70B & 39.3 & 90.0 & 76.7 & 34.9 & 95.6 & 40.7 & 62.9 \\
    InternLM2-20B-Chat & 20B & 37.1 & 87.8 & 76.7 & 34.5 & 92.2 & 30.4 & 59.8 \\
    CFGPT2-7B & 7B & \textbf{55.6} & 67.5 & 70.0 & 38.1 & 93.3 & 30.4 & 59.2 \\
    Yi1.5-34B-Chat & 34B & 35.9 & 93.3 & 73.2 & 34.5 & 86.7 & 22.3 & 57.7 \\
    ChatGLM3-6B & 6B & 35.9 & 67.5 & 76.7 & 31.4 & 86.7 & 42.8 & 56.8 \\
    DISC-FinLLM & 13B & 18.1 & 90.0 & 66.7 & 59.1 & 70.0 & 36.6 & 56.7 \\
    Qwen2.5-7B-Instruct & 7B & 37.1 & 87.8 & 76.7 & 31.4 & 90.0 & 15.0 & 56.3 \\
    XuanYuan2-70B-Chat & 70B & 33.6 & 66.6 & 83.3 & 31.2 & 78.9 & 25.4 & 53.2 \\
    Yi1.5-9B-Chat & 9B & 34.6 & 56.7 & 76.7 & 27.2 & 92.2 & 20.2 & 51.3 \\
    FinGPTv3.1 & 6B & 24.6 & 56.6 & 63.2 & 21.2 & 74.4 & 10.2 & 41.7 \\
    \bottomrule[2pt]
    \end{tabular}
}
\centering
\end{table*}

\begin{table*}[ht]
\caption{Evaluation Results (five-shot) for Finance Industry Knowledge (Average Similarity(\%)). These are objective short-answer questions including FTC: Financial Text Classification, FSA: Financial Sentiment Analysis, RE: Relation Extraction, FEE: Financial Event Extraction, NEE: Negative Entity Extraction, CEE: Causal Event Extraction.}
\label{FIK_obj_fiveshot}
\vspace{10pt}
\resizebox{0.8\textwidth}{!}{
    \begin{tabular}{llccccccc}
    \toprule[2pt]
    \textbf{Model} & \textbf{Size} & \textbf{FTC} & \textbf{FSA} & \textbf{RE} & \textbf{FEE} & \textbf{NEE} & \textbf{CEE} & \textbf{Average} \\
    \midrule[1pt]
    GPT-4o-mini & unknown & \textbf{55.7} & 90.0 & 72.2 & 78.5 & 90.0 & \textbf{69.7} & \textbf{76.0} \\
    Claude 3.5-sonnet & unknown & 54.5 & 81.1 & 81.1 & \textbf{79.8} & 92.2 & 65.7 & 75.7 \\ 
    Gemini-1.5-flash & unknown & 54.1 & \textbf{92.2} & 72.2 & 78.3 & 88.9 & 66.2 & 75.3 \\
    GPT-4o & unknown & 54.1 & 84.4 & 72.2 & 79.6 & 90.0 & 68.5 & 74.8 \\
    Gemini-1.5-pro & unknown & 47.0 & 86.7 & 67.8 & 75.0 & 83.3 & 65.1 & 70.8 \\
    Qwen2.5-72B-Instruct & 72B & 49.5 & 80.3 & 80.0 & 64.3 & 90.0 & 46.8 & 68.5 \\
    GLM4-9B-Chat & 9B & 37.1 & 76.7 & 77.8 & 66.1 & \textbf{96.7} & 54.5 & 68.2 \\
    Yi1.5-34B-Chat & 34B & 42.2 & 86.7 & 82.1 & 66.1 & 81.1 & 44.8 & 67.2 \\
    XuanYuan2-70B-Chat & 70B & 32.5 & 88.3 & 83.4 & 55.9 & 85.6 & 54.5 & 66.7 \\
    Baichuan2-13B-Chat & 13B & 34.5 & 76.7 & 78.3 & 75.3 & 83.3 & 50.8 & 66.5 \\
    Qwen2.5-7B-Instruct & 7B & 33.8 & 80.0 & 77.8 & 53.5 & 92.2 & 52.0 & 64.9 \\
    XuanYuan3-70B-Chat & 70B & 35.5 & 86.7 & 86.2 & 37.7 & 85.8 & 56.8 & 64.8 \\
    InternLM2-20B-Chat & 20B & 33.8 & 76.7 & 80.0 & 37.7 & 81.1 & 45.0 & 64.8 \\
    ChatGLM3-6B & 6B & 30.8 & 73.7 & 77.0 & 66.1 & 81.1 & 54.5 & 63.9 \\
    DISC-FinLLM & 13B & 25.0 & 61.1 & 53.3 & 53.9 & 96.7 & 50.8 & 56.6 \\
    InternLM2.5-20B-Chat & 20B & 32.5 & 62.2 & 83.4 & 47.7 & 48.4 & 60.7 & 55.8 \\
    Yi1.5-9B-Chat & 9B & 24.5 & 64.4 & \textbf{89.1} & 35.4 & 60.0 & 29.9 & 50.6 \\
    CFGPT2-7B & 7B & 53.5 & 38.9 & 67.8 & 26.4 & 66.7 & 45.0 & 49.7 \\
    FinGPTv3.1 & 6B & 16.5 & 30.9 & 59.8 & 18.4 & 52.0 & 21.9 & 33.3 \\
    \bottomrule[2pt]
    \end{tabular}
}
\centering
\end{table*}

\begin{table*}[ht]
\caption{Evaluation Results (zero-shot) for Finance Industry Knowledge (Average Similarity(\%)). These are subjective open-ended question including FTS: Financial Text Summarization, FCP: Financial Customer Portrait, MSR: Marketing Scripts Recommendation, IA: Investment Advice.}
\label{FIK_subj_zeroshot}
\vspace{10pt}
\resizebox{0.8\textwidth}{!}{
    \begin{tabular}{llccccc}
    \toprule[2pt]
    \textbf{Model} & \textbf{Size} & \textbf{FTS} & \textbf{FCP} & \textbf{MSR} & \textbf{IA} & \textbf{Average} \\
    \midrule[1pt]
    Qwen2.5-72B-Instruct & 72B & 31.0 & 80.0 & 22.2 & \textbf{24.0} & \textbf{39.3} \\
    Claude 3.5-sonnet & unknown & 25.9 & \textbf{83.3} & 22.2 & 23.9 & 38.8 \\
    Gemini-1.5-pro & unknown & 29.5 & 76.7 & 22.0 & 23.0 & 37.8 \\
    GPT-4o-mini & unknown & 28.6 & 76.7 & 22.0 & 23.7 & 37.8 \\
    InternLM2.5-20B-Chat & 20B & 27.9 & 76.7 & 22.1 & 23.9 & 37.7 \\
    Yi1.5-34B-Chat & 34B & 28.8 & 76.7 & \textbf{22.7} & 21.5 & 37.4 \\
    Gemini-1.5-flash & unknown & 30.1 & 73.3 & 22.2 & 23.4 & 37.3 \\
    XuanYuan2-70B-Chat & 70B & 28.5 & 73.3 & 22.2 & 23.3 & 36.8 \\
    CFGPT2-7B & 7B & \textbf{34.8} & 66.7 & 22.0 & 23.6 & 36.8 \\
    Qwen2.5-7B-Instruct & 7B & 26.4 & 73.3 & 22.1 & 23.9 & 36.4 \\
    ChatGLM3-6B & 6B & 29.0 & 71.3 & 22.1 & 23.3 & 36.4 \\
    InternLM2-20B-Chat & 20B & 28.8 & 70.0 & 22.1 & 23.5 & 36.1 \\
    GPT-4o & unknown & 28.4 & 70.0 & 22.0 & 23.7 & 36.0 \\
    GLM4-9B-Chat & 9B & 31.1 & 66.7 & 21.7 & 23.6 & 35.8 \\
    XuanYuan3-70B-Chat & 70B & 27.1 & 70.0 & 22.1 & 23.5 & 35.7 \\
    Yi1.5-9B-Chat & 9B & 32.6 & 60.0 & 22.6 & 23.4 & 34.7 \\
    DISC-FinLLM & 13B & 24.5 & 50.0 & 22.4 & 23.9 & 30.2 \\
    Baichuan2-13B-Chat & 13B & 27.2 & 46.7 & 22.0 & 23.8 & 29.9 \\
    FinGPTv3.1 & 6B & 22.5 & 43.2 & 22.0 & 22.5 & 27.6 \\
    \bottomrule[2pt]
    \end{tabular}
}
\centering
\end{table*}

\begin{table*}[ht]
\caption{Evaluation Results (five-shot) for Finance Industry Knowledge (Average Similarity(\%)). These are subjective open-ended question including FTS: Financial Text Summarization, FCP: Financial Customer Portrait, MSR: Marketing Scripts Recommendation, IA: Investment Advice.}
\label{FIK_subj_fiveshot}
\vspace{10pt}
\resizebox{0.8\textwidth}{!}{
    \begin{tabular}{llccccc}
    \toprule[2pt]
    \textbf{Model} & \textbf{Size} & \textbf{FTS} & \textbf{FCP} & \textbf{MSR} & \textbf{IA} & \textbf{Average} \\
    \midrule[1pt]
    XuanYuan3-70B-Chat & 70B & 35.2 & 83.3 & 22.4 & 23.8 & \textbf{41.2} \\
    Claude 3.5-sonnet & unknown & 35.0 & 83.3 & 22.2 & 23.7 & 41.1 \\
    Qwen2.5-7B-Instruct & 7B & 26.0 & \textbf{86.7} & \textbf{22.5} & \textbf{24.0} & 39.8 \\
    Qwen2.5-72B-Instruct & 72B & 31.6 & 80.0 & \textbf{22.5} & \textbf{24.0} & 39.5 \\
    XuanYuan2-70B-Chat & 70B & 31.6 & 80.0 & 22.4 & 23.8 & 39.5 \\
    Yi1.5-34B-Chat & 34B & \textbf{36.3} & 73.3 & 22.4 & 23.8 & 39.0 \\
    GLM4-9B-Chat & 9B & 35.0 & 73.3 & 22.4 & 23.8 & 38.6 \\
    Cemini-1.5-pro & unknown & 29.5 & 76.7 & 22.0 & 23.0 & 37.8 \\
    GPT-4o-mini & unknown & 28.6 & 76.7 & 22.0 & 23.7 & 37.8 \\
    Gemini-1.5-flash & unknown & 30.1 & 73.3 & 22.2 & 23.4 & 37.3 \\
    Baichuan2-13B-Chat & 13B & 35.9 & 66.7 & 22.2 & 23.5 & 37.1 \\
    ChatGLM3-6B & 6B & 25.8 & 73.3 & 22.3 & 23.8 & 36.3 \\
    GPT-4o & unknown & 28.4 & 70.0 & 22.0 & 23.7 & 36.0 \\
    InternLM2.5-20B-Chat & 20B & 31.8 & 56.7 & 22.3 & \textbf{24.0} & 33.7 \\
    InternLM2-20B-Chat & 20B & 31.6 & 56.7 & 22.4 & 23.8 & 33.6 \\
    Yi1.5-9B-Chat & 9B & 11.1 & 76.7 & 22.4 & 23.8 & 33.5 \\
    DISC-FinLLM & 13B & 33.3 & 46.7 & \textbf{22.5} & 23.8 & 31.6 \\
    CFGPT2-7B & 7B & 30.5 & 36.7 & 22.3 & \textbf{24.0} & 28.4 \\
    FinGPTv3.1 & 6B & 15.6 & 48.5 & 22.0 & 22.5 & 27.2 \\
    \bottomrule[2pt]
    \end{tabular}
}
\centering
\end{table*}

\begin{table*}[ht]
\caption{Evaluation Results (zero-shot CoT) for Finance Industry Knowledge (Average Similarity(\%)). These are objective short-answer questions including FTC: Financial Text Classification, FSA: Financial Sentiment Analysis, RE: Relation Extraction, FEE: Financial Event Extraction, NEE: Negative Entity Extraction, CEE: Causal Event Extraction.}
\label{FIK_obj_zeroshotCot}
\vspace{10pt}
\resizebox{0.8\textwidth}{!}{
    \begin{tabular}{llcccccc}
    \toprule[2pt]
    \textbf{Model} & \textbf{Size} & \textbf{FTC} & \textbf{RE} & \textbf{FEE} & \textbf{NEE} & \textbf{CEE} & \textbf{Average} \\
    \midrule[1pt]
    GPT-4o & unknown & \textbf{53.2} & \textbf{86.7} & 76.4 & 88.4 & \textbf{68.0} & \textbf{74.5} \\
    GPT-4o-mini & unknown & 52.4 & 83.3 & 74.5 & 90.0 & 62.0 & 72.4 \\
    Gemini-1.5-pro & unknown & 48.6 & 80.0 & 78.1 & 86.7 & 57.1 & 70.1 \\
    Claude 3.5-sonnet & unknown & 51.7 & 80.0 & 72.2 & 90.0 & 54.8 & 69.7 \\
    GLM4-9B-Chat & 9B & 42.0 & \textbf{86.7} & 58.6 & 96.7 & 57.9 & 68.4 \\
    Gemini-1.5-flash & unknown & 52.3 & 83.3 & \textbf{78.3} & 81.7 & 45.7 & 68.3 \\
    XuanYuan3-70B-Chat & 70B & 31.7 & 80.0 & 67.8 & 96.7 & 58.7 & 67.0 \\
    Yi1.5-9B-Chat & 9B & 40.4 & 66.7 & 71.5 & \textbf{100.0} & 55.3 & 66.8 \\
    Baichuan2-13B-Chat & 13B & 22.2 & 80.0 & 74.5 & 96.7 & 59.2 & 66.5 \\
    Yi1.5-34B-Chat & 34B & 37.0 & 73.3 & 66.4 & 96.7 & 54.6 & 65.6 \\
    InternLM2.5-20B-Chat & 20B & 45.0 & 80.0 & 53.6 & 93.3 & 53.5 & 65.1 \\
    Qwen2.5-72B-Instruct & 72B & 40.4 & 76.7 & 56.8 & 90.0 & 58.7 & 64.5 \\
    XuanYuan2-70B-Chat & 70B & 26.7 & 76.7 & 66.4 & 90.0 & 54.6 & 62.9 \\
    InternLM2-20B-Chat & 20B & 37.0 & 73.3 & 58.6 & 90.0 & 54.6 & 62.7 \\
    ChatGLM3-6B & 6B & 35.0 & 71.3 & 56.6 & 88.0 & 52.6 & 60.7 \\
    Qwen2.5-7B-Instruct & 7B & 26.7 & 73.3 & 57.9 & 90.0 & 51.5 & 59.9 \\
    CFGPT2-7B & 7B & 51.5 & 73.3 & 37.4 & 86.7 & 30.2 & 55.8 \\
    FinGPTv3.1 & 6B & 14.7 & 59.3 & 41.6 & 76.0 & 39.5 & 46.2 \\
    DISC-FinLLM & 13B & 26.7 & 63.3 & 37.4 & 76.7 & 19.9 & 44.8 \\
    \bottomrule[2pt]
    \end{tabular}
}
\centering
\end{table*}

\begin{table*}[ht]
\caption{Evaluation Results (five-shot CoT) for Finance Industry Knowledge (Average Similarity(\%)). These are objective short-answer questions including FTC: Financial Text Classification, FSA: Financial Sentiment Analysis, RE: Relation Extraction, FEE: Financial Event Extraction, NEE: Negative Entity Extraction, CEE: Causal Event Extraction.}
\label{FIK_obj_fiveshotCot}
\vspace{10pt}
\resizebox{0.8\textwidth}{!}{
    \begin{tabular}{llcccccc}
    \toprule[2pt]
    \textbf{Model} & \textbf{Size} & \textbf{FTC} & \textbf{RE} & \textbf{FEE} & \textbf{NEE} & \textbf{CEE} & \textbf{Average} \\
    \midrule[1pt]
    GLM4-9B-Chat & 9B & \textbf{61.2} & \textbf{83.3} & 60.1 & \textbf{96.7} & 65.4 & \textbf{73.3} \\
    Qwen2.5-72B-Instruct & 72B & 46.3 & \textbf{83.3} & 80.1 & 86.7 & 62.9 & 71.9 \\
    InternLM2.5-20B-Chat & 20B & 40.0 & 80.0 & 79.8 & 86.7 & \textbf{66.4} & 70.6 \\
    Gemini-1.5-flash & unknown & 52.4 & 70.0 & 75.5 & 89.6 & 65.6 & 70.6 \\
    GPT-4o-mini & unknown & 52.9 & 73.3 & 76.3 & 88.3 & 60.2 & 70.2 \\
    GPT-4o & unknown & 56.7 & 66.7 & 79.0 & 87.4 & 60.9 & 70.1 \\
    Yi1.5-9B-Chat & 9B & 50.0 & \textbf{83.3} & 76.8 & 73.3 & 62.6 & 69.2 \\
    Claude 3.5-sonnet & unknown & 47.7 & 76.7 & 71.0 & 87.4 & 62.9 & 69.1 \\
    Yi1.5-34B-Chat & 34B & 47.4 & 73.3 & \textbf{83.1} & 73.3 & 61.9 & 67.8 \\
    XuanYuan3-70B-Chat & 70B & 36.4 & \textbf{83.3} & 78.2 & 86.7 & 50.3 & 67.0 \\
    InternLM2-20B-Chat & 20B & 36.4 & 80.0 & 76.8 & 80.0 & 58.7 & 66.4 \\
    Qwen2.5-7B-Instruct & 7B & 28.9 & 80.0 & 73.2 & \textbf{96.7} & 51.3 & 66.0 \\
    XuanYuan2-70B-Chat & 70B & 36.4 & 76.7 & 78.2 & 73.3 & 58.7 & 64.7 \\
    gemini-1.5-pro & unknown & 43.8 & 66.7 & 76.0 & 78.0 & 57.1 & 64.3 \\
    ChatGLM3-6B & 6B & 33.4 & 77.0 & 73.8 & 77.0 & 55.7 & 63.4 \\
    CFGPT2-7B & 7B & 36.4 & 70.0 & 38.9 & 80.0 & 38.7 & 52.8 \\
    Baichuan2-13B-Chat & 13B & 22.2 & 66.7 & 26.3 & 73.3 & 44.8 & 46.7 \\
    DISC-FinLLM & 13B & 28.9 & 70.0 & 42.3 & 67.3 & 17.3 & 45.2 \\
    FinGPTv3.1 & 6B & 16.2 & 60.7 & 20.3 & 67.3 & 32.7 & 39.4 \\
    \bottomrule[2pt]
    \end{tabular}
}
\centering
\end{table*}

\begin{table*}[ht]
\caption{Evaluation Results (zero-shot CoT) for Finance Industry Knowledge (Average Similarity(\%)). These are subjective open-ended question including FTS: Financial Text Summarization, FCP: Financial Customer Portrait, IA: Investment Advice.}
\label{FIK_subj_zeroshotCot}
\vspace{10pt}
\resizebox{0.7\textwidth}{!}{
    \begin{tabular}{llcccc}
    \toprule[2pt]
    \textbf{Model} & \textbf{Size} & \textbf{FTS} & \textbf{FCP} & \textbf{IA} & \textbf{Average} \\
    \midrule[1pt]
    Qwen2.5-72B-Instruct & 72B & 29.0 & \textbf{83.3} & 22.8 & \textbf{45.0} \\
    InternLM2.5-20B-Chat & 20B & 28.2 & \textbf{83.3} & 23.2 & 44.9 \\
    GPT-4o-mini & unknown & 30.4 & 80.0 & 22.3 & 44.2 \\
    Claude 3.5-sonnet & unknown & 29.4 & 76.7 & 22.2 & 42.8 \\
    Gemini-1.5-pro & unknown & 31.7 & 73.3 & 22.2 & 42.4 \\
    GPT-4o & unknown & 29.3 & 73.3 & 23.0 & 41.9 \\
    Yi1.5-9B-Chat & 9B & 29.0 & 73.3 & 22.5 & 41.6 \\
    Qwen2.5-7B-Instruct & 7B & 24.3 & 76.7 & 23.0 & 41.3 \\
    GLM4-9B-Chat & 9B & 30.7 & 70.0 & 22.2 & 41.0 \\
    InternLM2-20B-Chat & 20B & 28.7 & 70.0 & 22.5 & 40.4 \\
    XuanYuan3-70B-Chat & 70B & 28.2 & 70.0 & 22.4 & 40.2 \\
    Gemini-1.5-flash & unknown & 31.4 & 66.7 & 21.8 & 40.0 \\
    XuanYuan2-70B-Chat & 70B & 24.3 & 70.0 & 23.0 & 39.1 \\
    ChatGLM3-6B & 6B & 24.2 & 70.0 & 22.1 & 38.8 \\
    CFGPT2-7B & 7B & \textbf{34.7} & 50.0 & 23.1 & 35.9 \\
    Yi1.5-34B-Chat & 34B & 28.7 & 56.7 & 22.1 & 35.8 \\
    DISC-FinLLM & 13B & 26.9 & 53.3 & \textbf{23.3} & 34.5 \\
    Baichuan2-13B-Chat & 13B & 20.1 & 53.3 & 23.2 & 32.2 \\
    FinGPTv3.1 & 6B & 18.2 & 44.2 & 22.2 & 28.2 \\
    \bottomrule[2pt]
    \end{tabular}
}
\centering
\end{table*}

\begin{table*}[ht]
\caption{Evaluation Results (five-shot CoT) for Finance Industry Knowledge (Average Similarity(\%)). These are subjective open-ended question including FTS: Financial Text Summarization, FCP: Financial Customer Portrait, IA: Investment Advice.}
\label{FIK_subj_fiveshotCot}
\vspace{10pt}
\resizebox{0.7\textwidth}{!}{
    \begin{tabular}{llcccc}
    \toprule[2pt]
    \textbf{Model} & \textbf{Size} & \textbf{FTS} & \textbf{FCP} & \textbf{IA} & \textbf{Average} \\
    \midrule[1pt]
    Claude 3.5-sonnet & unknown & 28.3 & \textbf{83.3} & 21.8 & \textbf{44.5} \\
    Yi1.5-9B-Chat & 9B & \textbf{35.8} & 73.3 & 22.2 & 43.8 \\
    Qwen2.5-72B-Instruct & 72B & 34.2 & 73.3 & 23.0 & 43.5 \\
    InternLM2.5-20B-Chat & 20B & 29.8 & 76.7 & 22.8 & 43.1 \\
    Gemini-1.5-pro & unknown & 24.4 & 80.0 & 22.9 & 42.4 \\
    GPT-4o & unknown & 27.8 & 73.3 & 22.2 & 41.1 \\
    XuanYuan3-70B-Chat & 70B & 25.6 & 73.3 & 23.0 & 40.6 \\
    GLM4-9B-Chat & 9B & 35.1 & 63.3 & 22.7 & 40.4 \\
    GPT-4o-mini & unknown & 27.8 & 70.0 & 22.8 & 40.2 \\
    XuanYuan2-70B-Chat & 70B & 24.3 & 73.3 & 22.2 & 39.9 \\
    Yi1.5-34B-Chat & 34B & 32.2 & 63.3 & 22.6 & 39.4 \\
    Gemini-1.5-flash & unknown & 28.3 & 66.7 & 22.7 & 39.2 \\
    Qwen2.5-7B-Instruct & 7B & 27.8 & 63.3 & 22.6 & 37.9 \\
    InternLM2-20B-Chat & 20B & 27.8 & 63.3 & 22.6 & 37.9 \\
    ChatGLM3-6B & 6B & 24.3 & 63.3 & 22.2 & 36.6 \\
    CFGPT2-7B & 7B & 33.6 & 53.3 & 22.6 & 36.5 \\
    Baichuan2-13B-Chat & 13B & 30.5 & 43.3 & 22.8 & 32.2 \\
    DISC-FinLLM & 13B & 26.5 & 45.6 & \textbf{23.3} & 31.8 \\
    FinGPTv3.1 & 6B & 22.5 & 35.9 & 22.1 & 26.8 \\
    \bottomrule[2pt]
    \end{tabular}
}
\centering
\end{table*}

\begin{table*}[ht]
\caption{Evaluation Results (zero-shot) for Finance Security Knowledge (Average Accuracy(\%)). App: Application security, Cryp: Cryptographic protection, MA: Malware analysis, MS: Memory security, NS: Network security, Pent: Pentest, Reve: Reverse engineering, Soft: Software security, Syst: System security, Vul: Vulnerability detection, WS: Web security}
\label{FSK_zeroshot}
\vspace{10pt}
\resizebox{1\textwidth}{!}{
    \begin{tabular}{llccccccccccccc}
    \toprule[2pt]
    \textbf{Model} & \textbf{Size} & \textbf{App} & \textbf{Cryp} & \textbf{MA} & \textbf{MS} & \textbf{NS} & \textbf{Pent} & \textbf{Reve} & \textbf{Soft} & \textbf{Syst} & \textbf{Vul} & \textbf{WS} & \textbf{Average} \\
    \midrule[1pt]
    GPT-4o & unknown & 77.8 & 70.4 & 77.8 & \textbf{92.6} & 70.4 & \textbf{96.3} & 85.2 & 81.5 & 85.2 & \textbf{81.5} & 81.5 &\textbf{81.8} \\
    Qwen2.5-72B-Instruct & 72B & 77.8 & \textbf{85.2} & \textbf{81.5} & 81.5 & 77.8 & 77.8 & 81.5 & 77.8 & \textbf{92.6} & \textbf{81.5} & 85.2 & 81.8 \\
    GPT-4o-mini & unknown & 74.1 & 70.4 & 74.1 & 85.2 & 77.8 & 88.9 & 77.8 & 81.5 & 81.5 & 77.8 & 81.5 & 79.1 \\
    Claude 3.5-sonnet & unknown & 70.4 & 66.7 & 70.4 & 81.5 & \textbf{81.5} & 85.2 & 85.2 & 81.5 & 81.5 & 70.4 & 85.2 & 78.1 \\
    Gemini-1.5-pro & unknown & 81.5 & 55.6 & 77.8 & 81.5 & 70.4 & 88.9 & 77.8 & \textbf{85.2} & 85.2 & 70.4 & 81.5 & 77.8 \\
    Gemini-1.5-flash & unknown & 77.8 & 59.3 & 77.8 & 85.2 & \textbf{81.5} & 88.9 & 85.2 & 70.4 & 81.5 & 70.4 & 74.1 & 77.5 \\
    Yi1.5-34B-Chat & 34B & 70.4 & 77.8 & 74.1 & 74.1 & 70.4 & 78.8 & 78.8 & 74.1 & 88.9 & 74.1 & 74.1 & 76.0 \\
    XuanYuan3-70B-Chat & 70B & 77.8 & 77.8 & 66.7 & 66.7 & 74.1 & 92.6 & 70.4 & 63.0 & 88.9 & 70.4 & 70.4 & 74.4 \\
    InternLM2.5-20B-Chat & 20B & \textbf{85.2} & 77.8 & 55.6 & 77.8 & 59.3 & 74.1 & 74.1 & 63.0 & 81.5 & 77.8 & \textbf{88.9} & 74.1 \\
    GLM4-9B-Chat & 9B & 66.7 & 77.8 & 55.6 & 70.4 & 74.1 & 77.8 & \textbf{88.9} & 66.7 & 81.5 & 66.7 & 77.8 & 73.1 \\
    InternLM2-20B-Chat & 20B & 74.1 & 77.8 & 63.0 & 70.4 & 70.4 & 77.8 & 74.1 & 66.7 & 81.5 & 70.4 & 77.8 & 73.1 \\
    Qwen2.5-7B-Instruct & 7B & 74.1 & 81.5 & 63.0 & 74.1 & 63.0 & 74.1 & 66.7 & 70.4 & 74.1 & 66.7 & 81.5 & 71.7 \\
    Yi1.5-9B-Chat & 9B & 63.0 & 77.8 & 70.4 & 70.4 & 74.1 & 66.7 & 85.2 & 66.7 & 77.8 & 55.6 & 77.8 & 71.4 \\
    XuanYuan2-70B-Chat & 70B & 74.1 & 74.1 & 55.6 & 44.4 & 59.3 & 66.7 & 74.1 & 59.3 & 88.9 & 74.1 & 77.8 & 68.0 \\
    CFGPT2-7B & 7B & 66.7 & 70.4 & 59.2 & 59.2 & 66.7 & 77.8 & 60.4 & 55.6 & 70.4 & 59.3 & 70.4 & 65.1 \\
    Baichuan2-13B-Chat & 13B & 66.7 & 66.7 & 40.7 & 51.9 & 63.0 & 63.0 & 70.4 & 55.6 & 70.4 & 63.0 & 66.7 & 61.6 \\
    ChatGLM3-6B & 6B & 44.4 & 55.6 & 40.7 & 29.6 & 55.6 & 51.9 & 59.3 & 33.3 & 59.3 & 40.7 & 59.3 & 48.2 \\
    DISC-FinLLM & 13B & 25.9 & 22.2 & 18.5 & 11.1 & 18.5 & 25.9 & 37.0 & 37.0 & 29.6 & 29.6 & 22.2 & 25.2 \\
    FinGPTv3.1 & 6B & 24.4 & 19.5 & 14.8 & 7.7 & 17.1 & 21.4 & 36.4 & 30.9 & 27.9 & 29.1 & 19.8 & 22.7 \\
    \bottomrule[2pt]
    \end{tabular}
}
\centering
\end{table*}

\begin{table*}[ht]
\caption{Evaluation Results (five-shot) for Finance Security Knowledge (Average Accuracy(\%)). App: Application security, Cryp: Cryptographic protection, MA: Malware analysis, MS: Memory security, NS: Network security, Pent: Pentest, Reve: Reverse engineering, Soft: Software security, Syst: System security, Vul: Vulnerability detection, WS: Web security}
\label{FSK_fiveshot}
\vspace{10pt}
\resizebox{1\textwidth}{!}{
    \begin{tabular}{llccccccccccccc}
    \toprule[2pt]
    \textbf{Model} & \textbf{Size} & \textbf{App} & \textbf{Cryp} & \textbf{MA} & \textbf{MS} & \textbf{NS} & \textbf{Pent} & \textbf{Reve} & \textbf{Soft} & \textbf{Syst} & \textbf{Vul} & \textbf{WS} & \textbf{Average} \\
    \midrule[1pt]
    GPT-4o & unknown & 74.1 & 70.4 & 74.1 & \textbf{92.6} & 81.5 & \textbf{96.3} & \textbf{81.5} & 88.9 & 85.2 & 74.1 & 81.5 & \textbf{81.8} \\
    GPT-4o-mini & unknown & 74.1 & 77.8 & 77.8 & 88.9 & 70.4 & 88.9 & \textbf{81.5} & \textbf{92.6} & 77.8 & \textbf{81.5} & 81.5 & 81.2 \\
    Qwen2.5-72B-Instruct & 72B & 74.1 & \textbf{81.5} & \textbf{85.2} & 74.1 & 70.4 & 85.2 & \textbf{81.5} & 77.8 & \textbf{92.6} & 74.1 & 85.2 & 80.2 \\
    Claude 3.5-sonnet & unknown & 66.7 & 70.4 & 70.4 & 88.9 & 77.8 & 92.6 & \textbf{81.5} & 85.2 & 81.5 & 74.1 & \textbf{88.9} & 79.8 \\
    Gemini-1.5-flash & unknown & 77.8 & 55.6 & \textbf{85.2} & 88.9 & \textbf{85.2} & \textbf{96.3} & 74.1 & 77.8 & 81.5 & 74.1 & 74.1 & 79.1 \\
    Gemini-1.5-pro & unknown & 81.5 & 63.0 & 74.1 & 81.5 & 66.7 & 85.2 & \textbf{81.5} & 88.9 & 81.5 & 66.7 & 74.1 & 76.8 \\
    InternLM2.5-20B-Chat & 20B & \textbf{85.2} & 74.1 & 70.4 & 70.4 & 74.1 & 70.4 & 70.1 & 74.1 & 81.5 & 77.8 & 77.8 & 75.1 \\
    Yi1.5-34B-Chat & 34B & 59.3 & 77.8 & 70.4 & 74.1 & 74.1 & 81.5 & 77.8 & 70.4 & 81.5 & 63.0 & 74.1 & 73.1 \\
    XuanYuan3-70B-Chat & 70B & 74.1 & \textbf{81.5} & 59.3 & 59.3 & 74.1 & 85.2 & \textbf{81.5} & 55.6 & 88.9 & 59.3 & 74.1 & 72.1 \\
    Qwen2.5-7B-Instruct & 7B & 66.7 & 66.7 & 51.9 & 74.1 & 63.0 & 77.8 & 81.4 & 70.3 & 81.4 & 77.8 & 74.1 & 71.4 \\
    InternLM2-20B-Chat & 20B & 70.3 & 78.2 & 55.6 & 66.7 & 74.1 & 81.5 & 79.5 & 55.6 & 81.5 & 66.7 & 74.1 & 71.3 \\
    XuanYuan2-70B-Chat & 70B & 65.2 & 78.2 & 55.5 & 57.6 & 74.1 & 82.5 & 79.5 & 54.6 & 84.2 & 65.2 & 74.1 & 70.1 \\
    CFGPT2-7B & 7B & 70.3 & \textbf{81.5} & 55.6 & 66.7 & 77.8 & 77.8 & 59.3 & 51.9 & 77.8 & 66.7 & 63.0 & 68.0 \\
    GLM4-9B-Chat & 9B & 70.4 & 74.1 & 37.0 & 63.0 & 37.0 & 51.9 & 66.7 & 48.2 & 81.5 & 70.4 & 55.6 & 59.6 \\
    ChatGLM3-6B & 6B & 48.1 & 59.3 & 37.0 & 37.0 & 55.6 & 55.6 & 51.9 & 51.9 & 55.6 & 51.9 & 59.3 & 51.2 \\
    Baichuan2-13B-Chat & 13B & 48.1 & 40.7 & 48.1 & 29.6 & 44.4 & 59.3 & 59.3 & 33.3 & 55.6 & 59.3 & 51.9 & 48.1 \\
    Yi1.5-9B-Chat & 9B & 51.9 & 55.6 & 37.0 & 40.7 & 44.4 & 59.3 & 33.3 & 37.0 & 51.9 & 51.9 & 44.4 & 46.1 \\
    DISC-FinLLM & 13B & 11.1 & 14.8 & 25.9 & 22.2 & 40.7 & 29.6 & 22.2 & 29.6 & 25.9 & 33.3 & 33.3 & 26.2 \\
    FinGPTv3.1 & 6B & 7.1 & 12.1 & 23.0 & 19.2 & 35.9 & 28.7 & 20.0 & 27.4 & 21.6 & 30.5 & 32.9 & 23.5 \\
    \bottomrule[2pt]
    \end{tabular}
}
\centering
\end{table*}

\begin{table*}[ht]
\caption{Evaluation Results (zero-shot CoT) for Finance Security Knowledge (Average Accuracy(\%)). App: Application security, Cryp: Cryptographic protection, MA: Malware analysis, MS: Memory security, NS: Network security, Pent: Pentest, Reve: Reverse engineering, Soft: Software security, Syst: System security, Vul: Vulnerability detection, WS: Web security}
\label{FSK_zeroshotCot}
\vspace{10pt}
\resizebox{1\textwidth}{!}{
    \begin{tabular}{llccccccccccccc}
    \toprule[2pt]
    \textbf{Model} & \textbf{Size} & \textbf{App} & \textbf{Cryp} & \textbf{MA} & \textbf{MS} & \textbf{NS} & \textbf{Pent} & \textbf{Reve} & \textbf{Soft} & \textbf{Syst} & \textbf{Vul} & \textbf{WS} & \textbf{Average} \\
    \midrule[1pt]
    GPT-4o & unknown & 77.8 & 70.4 & \textbf{77.8} & \textbf{92.6} & 70.4 & \textbf{96.3} & \textbf{85.2} & 81.5 & 85.2 & \textbf{81.5} & 81.5 & \textbf{81.8} \\
    GPT-4o-mini & unknown & 77.8 & 74.1 & 74.1 & 88.9 & 74.1 & 88.9 & 77.8 & \textbf{85.2} & 74.1 & 77.8 & 81.5 & 79.5 \\
    Claude 3.5-sonnet & unknown & 59.3 & 77.8 & 66.7 & 85.2 & 77.8 & 88.9 & 81.5 & 81.5 & 81.5 & 77.8 & 81.5 & 78.1 \\
    Gemini-1.5-pro & unknown & \textbf{81.5} & 55.6 & 77.8 & 81.5 & 70.4 & 88.9 & 77.8 & 85.2 & 85.2 & 70.4 & 77.8 & 77.4 \\
    Gemini-1.5-flash & unknown & 81.5 & 55.6 & 70.4 & 85.2 & \textbf{85.2} & 88.9 & 81.5 & 70.4 & 85.2 & 74.1 & 70.4 & 77.1 \\
    InternLM2.5-20B-Chat & 20B & 74.1 & \textbf{81.5} & 55.6 & 81.5 & 70.4 & 74.1 & 77.8 & 55.6 & 81.5 & 74.1 & 77.8 & 73.1 \\
    Qwen2.5-72B-Instruct & 72B & 63.0 & 77.8 & 66.7 & 63.0 & 74.1 & 70.4 & 77.8 & 63.0 & \textbf{92.6} & 66.7 & 85.2 & 72.8 \\
    GLM4-9B-Chat & 9B & 74.1 & 63.0 & 63.0 & 66.7 & 59.4 & 74.1 & 77.8 & 63.0 & 70.4 & 74.1 & \textbf{88.9} & 70.4 \\
    XuanYuan3-70B-Chat & 70B & 77.8 & 59.3 & 63.0 & 51.9 & 66.7 & 59.3 & 81.5 & 51.9 & 81.5 & 59.3 & 74.1 & 66.0 \\
    Yi1.5-9B-Chat & 9B & 59.3 & 77.8 & 70.4 & 74.1 & 55.6 & 51.9 & 66.7 & 59.3 & 74.1 & 63.0 & 70.4 & 65.7 \\
    XuanYuan2-70B-Chat & 70B & 76.2 & 59.3 & 50.2 & 47.6 & 65.2 & 59.3 & 81.5 & 64.1 & 69.3 & 50.2 & 74.1 & 63.4 \\
    InternLM2-20B-Chat & 20B & 66.7 & 59.3 & 63.0 & 51.9 & 59.4 & 59.3 & 77.8 & 59.3 & 70.4 & 59.3 & 70.4 & 63.3 \\
    CFGPT2-7B & 7B & 70.4 & 59.3 & 51.8 & 48.1 & 63.0 & 77.8 & 66.7 & 74.1 & 66.7 & 51.9 & 66.7 & 63.3 \\
    Yi1.5-34B-Chat & 34B & 66.7 & 70.4 & 63.0 & 59.3 & 55.6 & 63.0 & 66.7 & 48.1 & 74.1 & 63.0 & 63.0 & 63.0 \\
    Baichuan2-13B-Chat & 13B & 59.3 & 63.0 & 40.7 & 44.4 & 48.2 & 66.7 & 70.4 & 40.7 & 77.8 & 48.2 & 74.1 & 57.6 \\
    ChatGLM3-6B & 6B & 55.6 & 40.7 & 51.9 & 44.4 & 59.3 & 44.4 & 51.9 & 48.1 & 62.9 & 44.4 & 59.3 & 51.2 \\
    Qwen2.5-7B-Instruct & 7B & 55.6 & 55.6 & 51.9 & 51.9 & 33.3 & 51.9 & 51.9 & 51.9 & 44.4 & 40.7 & 65.2 & 50.4 \\
    DISC-FinLLM & 13B & 37.0 & 22.2 & 33.3 & 18.5 & 40.7 & 44.4 & 51.9 & 29.6 & 44.4 & 29.6 & 33.3 & 35.0 \\
    FinGPTv3.1 & 6B & 36.0 & 20.3 & 31.3 & 15.7 & 38.8 & 39.9 & 48.5 & 28.9 & 41.6 & 26.4 & 31.4 & 32.6 \\
    \bottomrule[2pt]
    \end{tabular}
}
\centering
\end{table*}

\begin{table*}[ht]
\caption{Evaluation Results (five-shot CoT) for Finance Security Knowledge (Average Accuracy(\%)). App: Application security, Cryp: Cryptographic protection, MA: Malware analysis, MS: Memory security, NS: Network security, Pent: Pentest, Reve: Reverse engineering, Soft: Software security, Syst: System security, Vul: Vulnerability detection, WS: Web security}
\label{FSK_fiveshotCot}
\vspace{10pt}
\resizebox{1\textwidth}{!}{
    \begin{tabular}{llccccccccccccc}
    \toprule[2pt]
    \textbf{Model} & \textbf{Size} & \textbf{App} & \textbf{Cryp} & \textbf{MA} & \textbf{MS} & \textbf{NS} & \textbf{Pent} & \textbf{Reve} & \textbf{Soft} & \textbf{Syst} & \textbf{Vul} & \textbf{WS} & \textbf{Average} \\
    \midrule[1pt]
    GPT-4o & unknown & 74.1 & 70.4 & 70.4 & \textbf{92.6} & \textbf{81.5} & \textbf{96.3} & 81.5 & \textbf{88.9} & 85.2 & 70.4 & 81.5 & \textbf{81.2} \\
    Gemini-1.5-flash & unknown & 77.8 & 59.3 & \textbf{85.2} & 88.9 & \textbf{81.5} & 92.6 & 77.8 & 81.5 & 85.2 & 66.7 & 70.4 & 78.8 \\
    GPT-4o-mini & unknown & 74.1 & 74.1 & 66.7 & 85.2 & 74.1 & 88.9 & 77.8 & \textbf{88.9} & 74.1 & \textbf{81.5} & 77.8 & 78.5 \\
    Qwen2.5-72B-Instruct & 72B & 74.1 & 74.1 & 77.8 & 81.5 & 77.8 & 66.7 & \textbf{85.2} & 81.5 & \textbf{92.6} & 66.7 & \textbf{85.2} & 78.5 \\
    Claude 3.5-sonnet & unknown & 66.7 & 59.3 & 63.0 & 85.2 & 74.1 & 92.6 & \textbf{85.2} & \textbf{88.9} & 81.5 & 74.1 & 81.5 & 77.4 \\
    Gemini-1.5-pro & unknown & 66.7 & 59.3 & 74.1 & 85.2 & 70.4 & 92.6 & 77.8 & 81.5 & 81.5 & 66.7 & 77.8 & 75.8 \\
    GLM4-9B-Chat & 9B & 63.0 & 63.0 & 66.7 & 63.0 & 59.3 & 74.1 & 81.5 & 63.0 & 77.8 & 63.0 & 70.4 & 67.7 \\
    InternLM2.5-20B-Chat & 20B & \textbf{85.2} & 59.3 & 66.7 & 70.4 & 55.6 & 51.9 & 51.9 & 59.3 & 70.4 & 63.0 & 74.1 & 64.3 \\
    XuanYuan3-70B-Chat & 70B & 77.8 & 51.9 & 70.4 & 55.6 & 63.0 & 55.6 & 81.5 & 51.9 & 70.4 & 55.6 & 63.0 & 63.3 \\
    Yi1.5-34B-Chat & 34B & 66.7 & 59.3 & 59.3 & 63.0 & 66.7 & 44.4 & 59.3 & 66.7 & 74.1 & 55.6 & 77.8 & 63.0 \\
    XuanYuan2-70B-Chat & 70B & 76.2 & 53.4 & 54.9 & 52.5 & 64.2 & 53.2 & 81.5 & 64.1 & 62.1 & 53.2 & 63.0 & 61.7 \\
    InternLM2-20B-Chat & 20B & 66.7 & 53.4 & 59.3 & 55.6 & 59.3 & 55.6 & 81.5 & 55.6 & 66.7 & 53.2 & 63.0 & 60.9 \\
    Yi1.5-9B-Chat & 9B & 55.6 & \textbf{77.8} & 59.3 & 55.6 & 55.6 & 44.4 & 51.9 & 55.6 & 63.0 & 51.9 & 85.2 & 59.6 \\
    CFGPT2-7B & 7B & 70.4 & 48.1 & 63.0 & 55.6 & 59.3 & 74.1 & 55.6 & 51.9 & 66.7 & 44.4 & 59.3 & 58.9 \\
    Baichuan2-13B-Chat & 13B & 55.6 & 51.9 & 40.7 & 37.0 & 48.2 & 48.2 & 55.6 & 37.0 & 55.6 & 40.7 & 55.6 & 47.8 \\
    ChatGLM3-6B & 6B & 40.7 & 55.6 & 51.9 & 37.0 & 48.1 & 55.6 & 44.4 & 40.7 & 55.6 & 40.7 & 51.8 & 47.5 \\
    Qwen2.5-7B-Instruct & 7B & 40.7 & 48.2 & 37.0 & 48.2 & 22.2 & 63.0 & 48.2 & 37.0 & 33.3 & 48.1 & 58.1 & 44.0 \\
    DISC-FinLLM & 13B & 33.3 & 25.9 & 29.6 & 29.6 & 29.6 & 44.4 & 25.9 & 29.6 & 44.4 & 18.5 & 29.6 & 30.9 \\
    FinGPTv3.1 & 6B & 27.1 & 22.4 & 22.6 & 25.4 & 22.0 & 38.2 & 22.4 & 26.9 & 43.1 & 18.4 & 26.9 & 26.9 \\
    \bottomrule[2pt]
    \end{tabular}
}
\centering
\end{table*}

\begin{table*}[t]
\caption{Evaluation Result (zero-shot) for Finance Agent(Similarity(\%)). COT: Chain of Thought, RAG: Retrieval Augmented Generation, FT: Financial tasks, MC: Multi-turn conversation, MD: Multi-document question and answer, API-I: API invocation, API-R: API retrieval}
\label{Agent_zeroshot}
\vspace{10pt}
\resizebox{1\textwidth}{!}{
    \begin{tabular}{llccccccccc}
    \toprule[2pt]
    \textbf{Model} & \textbf{Size} & \textbf{COT} & \textbf{RAG} & \textbf{FT} & \textbf{MC} & \textbf{MD} & \textbf{API-I} & \textbf{API-R} & \textbf{Average} \\
    \midrule[1pt]
    Claude 3.5-sonnet & unknown & \textbf{69.6} & \textbf{91.8} & \textbf{80.0} & \textbf{71.2} & \textbf{74.4} & \textbf{84.6} & \textbf{83.5} & \textbf{79.3} \\
    GPT-4o & unknown & 68.8 & 83.4 & \textbf{80.0} & 63.0 & 69.8 & 76.5 & 75.6 & 73.9 \\
    GPT-4o-mini & unknown & 63.0 & 82.6 & \textbf{80.0} & 63.0 & 68.9 & 77.4 & 75.2 & 72.9 \\
    Gemini-1.5-pro & unknown & 63.0 & 82.2 & \textbf{80.0} & 63.0 & 72.0 & 73.1 & 76.3 & 72.8 \\
    Gemini-1.5-flash & unknown & 50.4 & 85.2 & \textbf{80.0} & 62.2 & 71.1 & 72.2 & 75.2 & 70.9 \\
    Qwen2.5-72B-Instruct & 72B & 62.2 & 91.1 & 74.3 & 31.1 & 69.8 & 75.7 & 74.4 & 68.4 \\
    Qwen2.5-7B-Instruct & 7B & 44.4 & 86.7 & 77.8 & 63.0 & 66.3 & 59.6 & 68.9 & 66.7 \\
    Yi1.5-34B-Chat & 34B & 50.2 & 69.3 & 79.3 & 65.2 & 66.2 & 61.2 & 70.4 & 66.0 \\
    XuanYuan3-70B-Chat & 70B & 48.9 & 80.7 & 77.1 & 31.5 & 69.3 & 65.4 & 74.1 & 63.9 \\
    InternLM2.5-20B-Chat & 20B & 52.6 & 85.9 & \textbf{80.0} & 31.5 & 66.7 & 56.9 & 68.1 & 63.1 \\
    XuanYuan2-70B-Chat & 70B & 44.4 & 81.5 & 79.3 & 31.5 & 67.8 & 58.0 & 69.6 & 61.7 \\
    Yi1.5-9B-Chat & 9B & 45.2 & 66.0 & 69.6 & 63.7 & 65.7 & 50.2 & 67.0 & 61.1 \\
    InternLM2-20B-Chat & 20B & 40.2 & 75.7 & 72.1 & 58.0 & 61.3 & 54.6 & 64.6 & 60.9 \\
    GLM4-9B-Chat & 9B & 48.1 & 64.4 & 75.7 & 31.1 & 66.5 & 64.1 & 71.5 & 60.2 \\
    Baichuan2-13B-Chat & 13B & 31.9 & 67.4 & \textbf{80.0} & 31.5 & 66.1 & 46.5 & 66.3 & 55.7 \\
    CFGPT2-7B & 7B & 28.2 & 63.0 & 64.4 & 63.8 & 60.6 & 24.3 & 52.2 & 50.9 \\
    ChatGLM3-6B & 6B & 25.2 & 71.1 & 70.0 & 31.5 & 59.1 & 33.3 & 57.2 & 49.6 \\
    DISC-FinLLM & 13B & 20.7 & 61.5 & 62.9 & 31.1 & 43.1 & 21.3 & 52.0 & 41.8 \\
    FinGPTv3.1 & 6B & 4.6 & 48.2 & 42.8 & 31.1 & 36.4 & 25.3 & 30.0 & 31.2 \\
    \bottomrule[2pt]
    \end{tabular}
}
\centering
\end{table*}


\end{document}